\documentclass{article}

    \usepackage[preprint]{neurips_2019}

\usepackage[utf8]{inputenc} 
\usepackage[T1]{fontenc}    
\usepackage{hyperref}       
\usepackage{url}            
\usepackage{booktabs}       
\usepackage{amsfonts}       
\usepackage{amsmath}
\usepackage{nicefrac}       
\usepackage{microtype}      
\usepackage[pdftex]{graphicx}
\usepackage{subcaption}
\usepackage{float}
\newcommand*{\inlineequation}[2][]{%
  \begingroup
    \refstepcounter{equation}%
    \ifx\\#1\\%
    \else
      \label{#1}%
    \fi
    \relpenalty=10000 %
    \binoppenalty=10000 %
    \ensuremath{%
      #2%
    }%
    ~\@eqnnum
  \endgroup
}
\usepackage{todonotes}

\title{Recurrent Neural Processes}

\author{%
  Timon Willi \\
  NNAISENSE SA\\
  \texttt{timon@nnaisense.com}\\
  \And
  Jonathan Masci \\
  NNAISENSE SA\\
  \texttt{jonathan@nnaisense.com}\\
  \And
  Jürgen Schmidhuber\\
  NNAISENSE SA\\
  \texttt{juergen@nnaisense.com}\\
  \And
  Christian Osendorfer \\
  NNAISENSE SA \\
  \texttt{christian@nnaisense.com}\\
}

\begin{document}

\maketitle

\begin{abstract}
We extend Neural Processes (NPs) to sequential data through Recurrent NPs or RNPs, a family of conditional state space models. RNPs model the state space with Neural Processes. Given time series observed on fast real-world time scales but containing slow long-term variabilities, RNPs may derive appropriate slow latent time scales. They do so in an efficient manner by establishing conditional independence among subsequences of the time series. Our theoretically grounded framework for stochastic processes expands the applicability of NPs while retaining their benefits of flexibility, uncertainty estimation, and favorable runtime with respect to Gaussian Processes (GPs). We demonstrate that state spaces learned by RNPs benefit predictive performance on real-world time-series data and nonlinear system identification, even in the case of limited data availability.
\end{abstract}

\section{Introduction}
The surge of neural latent variable models for performing inference on stochastic processes is a recent development in Deep Learning. Neural Processes (NPs) \citep{Garnelo2018a} combine the best of neural networks and GPs, estimating uncertainty and model distributions over functions, while still achieving $\mathcal{O}(n+m)$ runtime instead of $\mathcal{O}\left((n+m)^{3}\right)$, where $n$ is the number of context points and $m$ the number of target points \citep{Garnelo2018}.\\
Our novel Recurrent NPs (RNPs) transfer the benefits of NPs to deep generative state-space models by modeling the state space with NPs. By introducing a hierarchy of NPs, the RNPs learn in the space-time domain and integrate information in space and time into a general model.\\
By modeling the state space with Neural Processes, we can capture multiple time scales.
Consider a time series of 10 years of hourly temperature measurements. Daily cycles (warm days, cold nights) are superimposed by seasonal cycles (winter, summer). RNPs can derive slow latent time scales (reflecting winter and summer) despite fast real-world time scales (reflecting hourly observations). RNPs retain the benefits from NPs of (a) estimating uncertainty, (b) modeling distributions over functions, and (c) high flexibility at test time without increasing runtime. RNPs enable efficient inference of the latent space by establishing conditional independence among subsequences in a given time series. We apply RNPs to various real-world one-step-look-ahead prediction tasks, illustrating their performance and benefits.\\
\section{Background}
 The name "Neural Process" has two components. The "Process" comes from the fact that we are learning a stochastic process (SP). A stochastic process $\{Z_{t}\}_{t \in T}$ is defined as a collection of random variables on a probability space $(\Omega, \mathcal{F},P)$ \citep{lamperti2012stochastic}. The random variables $Z_{t}$ are indexed by some set $T$. The random variables $Z_{t}$ take values in a common measure space $(\Xi,\mathcal{Z})$. Consider that $Z_{t}(\omega)$ takes two arguments: $Z(t,\omega)$, where $\omega \in \Omega$. If we fix $t$, we get the standard random variable $Z_{t}(\omega)$ from before. If we fix $\omega$, $Z(t)$ becomes a function from $T$ to $\Xi$. Basically, a stochastic process is a function-valued random variable. That is why the realization of a stochastic process is called a sample function $ Z: T \rightarrow \Xi$.\\
The "Neural" in Neural Processes comes from the fact that we are learning the SP with a Neural Network. To learn a stochastic process means to learn a distribution over functions mapping from an input $x\in \mathbb{R}^{d_{x}}$ to a random variable $Y \in \mathbb{R}^{d_{y}}$. To learn the distribution over functions, the NPs are cast as conditional latent variable models trying to learn the conditional probability distribution 
\begin{equation}
    P(Y_{T} | X_{T}, C)=\int P(Y_{T} | X_{T}, z) P(z | C) \mathrm{d} z\label{eq:cond_lat_mod}
\end{equation}
where $C=\left(X_{C}, Y_{C}\right)=\left(x_{i}, y_{i}\right)_{i \in \mathtt{I}(C)}$ are the context and $D=(X_{T}, Y_{T})=\left(x_{i}, y_{i}\right)_{i \in \mathtt{I}(D)}$ are the target points, and $z$ is some latent variable assumed to represent the parameterization of the stochastic process $\mathtt{P}: \{y_{x}\}_{x\in X}$. The $\mathtt{I}(\cdot)$ is returning the desired indices of a dataset. The points in $C$ and $D$ are assumed to come from the true stochastic process $\mathtt{P}$.\\
\section{Recurrent Neural Processes}
Next up we want to derive the Recurrent Neural Processes. The first step is to introduce time-dependent variants of the components of the NP. Therefore we define a time-dependent context $C_{t}$, where $t$ is the time step. The observations in $C_{t}$, $D_{t}$ are from a stochastic process $\mathtt{P}_{t}$, instead of only $\mathtt{P}$. In other words, we impose a temporal structure $\mathtt{P}_{t-1} \rightarrow \mathtt{P}_{t}$ with the underlying assumption that at each new timestep, there is a different stochastic process. Equivalently to the NPs, the processes $\mathtt{P}_{t}$ mentioned above are represented as distributions on latent variables $z_{t}$.
The sequential nature of the data leads to the assumption that it is helpful to make the current $z_{t}$ dependent, not only on the current context $C_{t}$, but also on previous timesteps $z_{<t}$, resulting in $P\left(z_{t} | z_{<t}, C_{t}\right)$. In order to access information from the previous step, we need to maintain a recurrent state throughout the sequence. Trivially the observation model is $P\left(\bold{Y}_{t} | \bold{X}_{t}, z_{t}\right)$. The bold letters indicate that the variables are observable. From here it follows that the generative process of the new model is
\begin{equation}
    P(\bold{Y}, Z | \bold{X}, C)=\prod_{t=1}^{T} P\left(\bold{Y}_{t} | \bold{X}_{t}, z_{t}\right) P\left(z_{t} | z_{<t}, C_{t}\right)\label{eq:snp}
\end{equation}
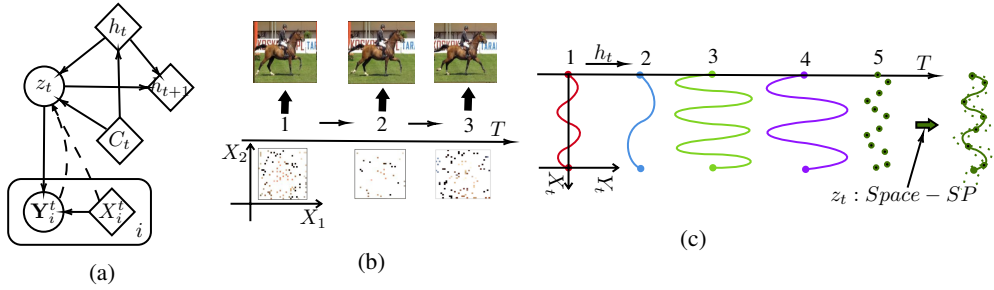
\begin{figure}[]
\begin{subfigure}{.2\textwidth}
     \centering
    \tikzset{every picture/.style={line width=0.75pt}} 
    \begin{tikzpicture}[x=0.75pt,y=0.75pt,yscale=-1,xscale=1,scale=0.4,every node/.style={scale=0.8}]
    
    \draw   (68.25,313.5) .. controls (68.25,299.69) and (79.44,288.5) .. (93.25,288.5) .. controls (107.06,288.5) and (118.25,299.69) .. (118.25,313.5) .. controls (118.25,327.31) and (107.06,338.5) .. (93.25,338.5) .. controls (79.44,338.5) and (68.25,327.31) .. (68.25,313.5) -- cycle ;
    \draw   (179.25,284) -- (208,313.5) -- (179.25,343) -- (150.5,313.5) -- cycle ;
    \draw   (48.5,288.6) .. controls (48.5,279.43) and (55.93,272) .. (65.1,272) -- (211.9,272) .. controls (221.07,272) and (228.5,279.43) .. (228.5,288.6) -- (228.5,338.4) .. controls (228.5,347.57) and (221.07,355) .. (211.9,355) -- (65.1,355) .. controls (55.93,355) and (48.5,347.57) .. (48.5,338.4) -- cycle ;
    \draw   (69,155) .. controls (69,141.19) and (80.19,130) .. (94,130) .. controls (107.81,130) and (119,141.19) .. (119,155) .. controls (119,168.81) and (107.81,180) .. (94,180) .. controls (80.19,180) and (69,168.81) .. (69,155) -- cycle ;
    \draw   (189.25,194) -- (218,223.5) -- (189.25,253) -- (160.5,223.5) -- cycle ;
    \draw   (187.25,51) -- (216,80.5) -- (187.25,110) -- (158.5,80.5) -- cycle ;
    \draw   (253.25,127) -- (282,156.5) -- (253.25,186) -- (224.5,156.5) -- cycle ;
    \draw    (94,180) -- (93.26,286.5) ;
    \draw [shift={(93.25,288.5)}, rotate = 270.4] [color={rgb, 255:red, 0; green, 0; blue, 0 }  ][line width=0.75]    (10.93,-3.29) .. controls (6.95,-1.4) and (3.31,-0.3) .. (0,0) .. controls (3.31,0.3) and (6.95,1.4) .. (10.93,3.29)   ;
    
    \draw    (171.5,93) -- (115.07,137.76) ;
    \draw [shift={(113.5,139)}, rotate = 321.58000000000004] [color={rgb, 255:red, 0; green, 0; blue, 0 }  ][line width=0.75]    (10.93,-3.29) .. controls (6.95,-1.4) and (3.31,-0.3) .. (0,0) .. controls (3.31,0.3) and (6.95,1.4) .. (10.93,3.29)   ;
    
    \draw    (172.5,209) -- (115.19,173.06) ;
    \draw [shift={(113.5,172)}, rotate = 392.09000000000003] [color={rgb, 255:red, 0; green, 0; blue, 0 }  ][line width=0.75]    (10.93,-3.29) .. controls (6.95,-1.4) and (3.31,-0.3) .. (0,0) .. controls (3.31,0.3) and (6.95,1.4) .. (10.93,3.29)   ;
    
    \draw    (222.5,156.47) -- (119,155) ;
    
    \draw [shift={(224.5,156.5)}, rotate = 180.81] [color={rgb, 255:red, 0; green, 0; blue, 0 }  ][line width=0.75]    (10.93,-3.29) .. controls (6.95,-1.4) and (3.31,-0.3) .. (0,0) .. controls (3.31,0.3) and (6.95,1.4) .. (10.93,3.29)   ;
    \draw    (189.25,194) -- (187.3,112) ;
    \draw [shift={(187.25,110)}, rotate = 448.64] [color={rgb, 255:red, 0; green, 0; blue, 0 }  ][line width=0.75]    (10.93,-3.29) .. controls (6.95,-1.4) and (3.31,-0.3) .. (0,0) .. controls (3.31,0.3) and (6.95,1.4) .. (10.93,3.29)   ;
    
    \draw    (202.5,96) -- (238.23,139.46) ;
    \draw [shift={(239.5,141)}, rotate = 230.57] [color={rgb, 255:red, 0; green, 0; blue, 0 }  ][line width=0.75]    (10.93,-3.29) .. controls (6.95,-1.4) and (3.31,-0.3) .. (0,0) .. controls (3.31,0.3) and (6.95,1.4) .. (10.93,3.29)   ;
    
    \draw    (150.5,313.5) -- (120.25,313.5) ;
    \draw [shift={(118.25,313.5)}, rotate = 360] [color={rgb, 255:red, 0; green, 0; blue, 0 }  ][line width=0.75]    (10.93,-3.29) .. controls (6.95,-1.4) and (3.31,-0.3) .. (0,0) .. controls (3.31,0.3) and (6.95,1.4) .. (10.93,3.29)   ;
    
    \draw  [dash pattern={on 4.5pt off 4.5pt}]  (113.5,296) .. controls (127.86,260.36) and (125.55,220.8) .. (104.15,178.29) ;
    \draw [shift={(103.5,177)}, rotate = 422.9] [fill={rgb, 255:red, 0; green, 0; blue, 0 }  ][line width=0.75]  [draw opacity=0] (8.93,-4.29) -- (0,0) -- (8.93,4.29) -- cycle    ;
    
    \draw  [dash pattern={on 4.5pt off 4.5pt}]  (163.5,300) .. controls (149.64,258.42) and (125.98,220.76) .. (104.16,178.29) ;
    \draw [shift={(103.5,177)}, rotate = 422.9] [fill={rgb, 255:red, 0; green, 0; blue, 0 }  ][line width=0.75]  [draw opacity=0] (8.93,-4.29) -- (0,0) -- (8.93,4.29) -- cycle    ;

    \draw (93.25,313.5) node   {$\mathbf{Y}^{t}_{i}$};
    \draw (179.25,313.5) node   {$X^{t}_{i}$};
    \draw (214.25,336.5) node   {$i$};
    \draw (94,155) node   {$z_{t}$};
    \draw (189.25,223.5) node   {$C_{t}$};
    \draw (253.25,156.5) node   {$h_{t+1}$};
    \draw (187.25,80.5) node   {$h_{t}$};

    \end{tikzpicture}
    \caption{\label{fig:snp_graphical_model}}
\end{subfigure}
\begin{subfigure}{.3\textwidth}
    \tikzset{every picture/.style={line width=0.75pt}} 
    \begin{tikzpicture}[x=0.75pt,y=0.75pt,yscale=-1,xscale=1,scale=0.4,every node/.style={scale=0.8}]
    
    \draw    (150.5,274) -- (489.5,275.99) ;
    \draw [shift={(491.5,276)}, rotate = 180.34] [color={rgb, 255:red, 0; green, 0; blue, 0 }  ][line width=0.75]    (10.93,-3.29) .. controls (6.95,-1.4) and (3.31,-0.3) .. (0,0) .. controls (3.31,0.3) and (6.95,1.4) .. (10.93,3.29)   ;
    
    \draw  [fill={rgb, 255:red, 0; green, 0; blue, 0 }  ,fill opacity=1 ] (432.35,235.14) -- (432.83,219.24) -- (429.33,219.14) -- (436.65,208.76) -- (443.33,219.57) -- (439.83,219.46) -- (439.34,235.35) -- cycle ;
    \draw  [fill={rgb, 255:red, 0; green, 0; blue, 0 }  ,fill opacity=1 ] (324.35,236.14) -- (324.83,220.24) -- (321.33,220.14) -- (328.65,209.76) -- (335.33,220.57) -- (331.83,220.46) -- (331.34,236.35) -- cycle ;
    \draw  [fill={rgb, 255:red, 0; green, 0; blue, 0 }  ,fill opacity=1 ] (201.35,239.14) -- (201.83,223.24) -- (198.33,223.14) -- (205.65,212.76) -- (212.33,223.57) -- (208.83,223.46) -- (208.34,239.35) -- cycle ;
    \draw    (248.5,252) -- (281.5,252) ;
    \draw [shift={(283.5,252)}, rotate = 180] [color={rgb, 255:red, 0; green, 0; blue, 0 }  ][line width=0.75]    (10.93,-3.29) .. controls (6.95,-1.4) and (3.31,-0.3) .. (0,0) .. controls (3.31,0.3) and (6.95,1.4) .. (10.93,3.29)   ;
    
    \draw    (360.5,253) -- (393.5,253) ;
    \draw [shift={(395.5,253)}, rotate = 180] [color={rgb, 255:red, 0; green, 0; blue, 0 }  ][line width=0.75]    (10.93,-3.29) .. controls (6.95,-1.4) and (3.31,-0.3) .. (0,0) .. controls (3.31,0.3) and (6.95,1.4) .. (10.93,3.29)   ;
    
    \draw (201.75,316.5) node  {\includegraphics[width=23.63pt,height=24.75pt]{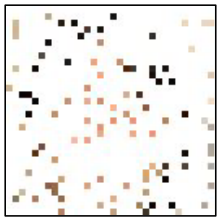}};
    \draw (324.75,316) node  {\includegraphics[width=25.13pt,height=24pt]{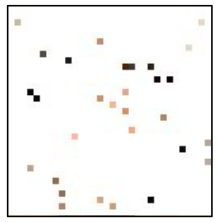}};
    \draw (428.75,316.5) node  {\includegraphics[width=28.13pt,height=24.75pt]{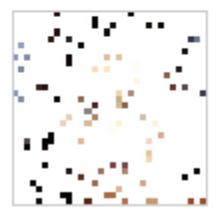}};
    \draw (204.75,162.5) node  {\includegraphics[width=29.63pt,height=29.25pt]{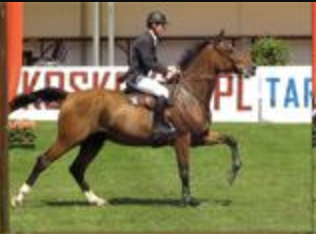}};
    \draw (326.75,163.5) node  {\includegraphics[width=31.13pt,height=29.25pt]{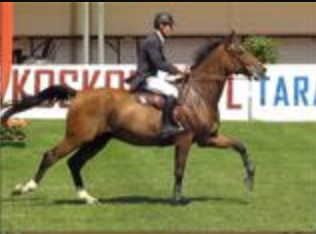}};
    \draw (432.75,163.5) node  {\includegraphics[width=29.63pt,height=27.75pt]{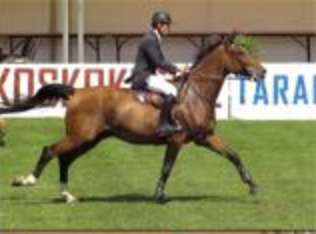}};
    \draw  (136.5,354) -- (251,354)(161.5,279) -- (161.5,376) (244,349) -- (251,354) -- (244,359) (156.5,286) -- (161.5,279) -- (166.5,286)  ;
    
    \draw (205.16,253.45) node  [align=left] {1};
    \draw (327.53,253.45) node  [align=left] {2};
    \draw (437.41,254.45) node  [align=left] {3};
    \draw (474,260) node   {$T$};
    \draw (241,373) node   {$X_{1}$};
    \draw (142,290) node   {$X_{2}$};

    \end{tikzpicture}
    \caption{\label{fig:snp_video}}
\end{subfigure}
\begin{subfigure}{.3\textwidth}
    \centering
    \tikzset{every picture/.style={line width=0.75pt}} 
    \begin{tikzpicture}[x=0.75pt,y=0.75pt,yscale=-1,xscale=1,scale=0.4,every node/.style={scale=0.8}]
    
    \draw [color={rgb, 255:red, 126; green, 211; blue, 33 }  ,draw opacity=1 ]   (247.5,58) .. controls (185.5,63) and (186.5,75) .. (248.5,79) .. controls (312.5,80) and (315.5,97) .. (248.5,99) .. controls (185.5,101) and (184.5,115) .. (248.5,121) .. controls (309.5,123) and (310.5,133) .. (249.5,140) .. controls (188.5,144) and (186.5,159) .. (249.5,159) .. controls (312.5,160) and (311.5,177) .. (248.5,176) ;
    \draw [shift={(248.5,176)}, rotate = 180.91] [color={rgb, 255:red, 126; green, 211; blue, 33 }  ,draw opacity=1 ][fill={rgb, 255:red, 126; green, 211; blue, 33 }  ,fill opacity=1 ][line width=0.75]      (0, 0) circle [x radius= 3.35, y radius= 3.35]   ;
    \draw [shift={(247.5,58)}, rotate = 175.39] [color={rgb, 255:red, 126; green, 211; blue, 33 }  ,draw opacity=1 ][fill={rgb, 255:red, 126; green, 211; blue, 33 }  ,fill opacity=1 ][line width=0.75]      (0, 0) circle [x radius= 3.35, y radius= 3.35]   ;
    \draw [color={rgb, 255:red, 208; green, 2; blue, 27 }  ,draw opacity=1 ]   (66,58.25) .. controls (82.5,71) and (81.5,78) .. (67.5,91) .. controls (49.5,101) and (45.5,109) .. (67.5,120) .. controls (87.5,127) and (83.5,139) .. (66.5,150) .. controls (42.5,159) and (56.5,175) .. (66.5,176) ;
    \draw [shift={(66.5,176)}, rotate = 5.71] [color={rgb, 255:red, 208; green, 2; blue, 27 }  ,draw opacity=1 ][fill={rgb, 255:red, 208; green, 2; blue, 27 }  ,fill opacity=1 ][line width=0.75]      (0, 0) circle [x radius= 3.35, y radius= 3.35]   ;
    \draw [shift={(66,58.25)}, rotate = 37.7] [color={rgb, 255:red, 208; green, 2; blue, 27 }  ,draw opacity=1 ][fill={rgb, 255:red, 208; green, 2; blue, 27 }  ,fill opacity=1 ][line width=0.75]      (0, 0) circle [x radius= 3.35, y radius= 3.35]   ;
    \draw [color={rgb, 255:red, 74; green, 144; blue, 226 }  ,draw opacity=1 ]   (156.5,59) .. controls (177.5,76) and (182.5,107) .. (160.5,117) .. controls (133.5,128) and (137.5,169) .. (157.5,177) ;
    \draw [shift={(157.5,177)}, rotate = 21.8] [color={rgb, 255:red, 74; green, 144; blue, 226 }  ,draw opacity=1 ][fill={rgb, 255:red, 74; green, 144; blue, 226 }  ,fill opacity=1 ][line width=0.75]      (0, 0) circle [x radius= 3.35, y radius= 3.35]   ;
    \draw [shift={(156.5,59)}, rotate = 38.99] [color={rgb, 255:red, 74; green, 144; blue, 226 }  ,draw opacity=1 ][fill={rgb, 255:red, 74; green, 144; blue, 226 }  ,fill opacity=1 ][line width=0.75]      (0, 0) circle [x radius= 3.35, y radius= 3.35]   ;
    \draw [color={rgb, 255:red, 144; green, 19; blue, 254 }  ,draw opacity=1 ]   (363.5,59) .. controls (309.5,66) and (292.5,84) .. (363.5,88) .. controls (438.5,94) and (433.5,113) .. (365.5,118) .. controls (302.5,125) and (299.5,138) .. (366.5,145) .. controls (439.5,157) and (430.5,171) .. (365.5,178) ;
    \draw [shift={(365.5,178)}, rotate = 173.85] [color={rgb, 255:red, 144; green, 19; blue, 254 }  ,draw opacity=1 ][fill={rgb, 255:red, 144; green, 19; blue, 254 }  ,fill opacity=1 ][line width=0.75]      (0, 0) circle [x radius= 3.35, y radius= 3.35]   ;
    \draw [shift={(363.5,59)}, rotate = 172.61] [color={rgb, 255:red, 144; green, 19; blue, 254 }  ,draw opacity=1 ][fill={rgb, 255:red, 144; green, 19; blue, 254 }  ,fill opacity=1 ][line width=0.75]      (0, 0) circle [x radius= 3.35, y radius= 3.35]   ;
    \draw    (27,58) -- (528.5,59.99) ;
    \draw [shift={(530.5,60)}, rotate = 180.23] [color={rgb, 255:red, 0; green, 0; blue, 0 }  ][line width=0.75]    (10.93,-3.29) .. controls (6.95,-1.4) and (3.31,-0.3) .. (0,0) .. controls (3.31,0.3) and (6.95,1.4) .. (10.93,3.29)   ;
    
    \draw  [color={rgb, 255:red, 65; green, 117; blue, 5 }  ,draw opacity=1 ][fill={rgb, 255:red, 0; green, 0; blue, 0 }  ,fill opacity=1 ] (453.5,59) .. controls (453.35,57.44) and (454.49,56.05) .. (456.06,55.9) .. controls (457.62,55.75) and (459.01,56.89) .. (459.16,58.46) .. controls (459.31,60.02) and (458.16,61.41) .. (456.6,61.56) .. controls (455.04,61.71) and (453.65,60.56) .. (453.5,59) -- cycle ;
    \draw  [color={rgb, 255:red, 65; green, 117; blue, 5 }  ,draw opacity=1 ][fill={rgb, 255:red, 0; green, 0; blue, 0 }  ,fill opacity=1 ] (466.5,68) .. controls (466.35,66.44) and (467.49,65.05) .. (469.06,64.9) .. controls (470.62,64.75) and (472.01,65.89) .. (472.16,67.46) .. controls (472.31,69.02) and (471.16,70.41) .. (469.6,70.56) .. controls (468.04,70.71) and (466.65,69.56) .. (466.5,68) -- cycle ;
    \draw  [color={rgb, 255:red, 65; green, 117; blue, 5 }  ,draw opacity=1 ][fill={rgb, 255:red, 0; green, 0; blue, 0 }  ,fill opacity=1 ] (455.5,82) .. controls (455.35,80.44) and (456.49,79.05) .. (458.06,78.9) .. controls (459.62,78.75) and (461.01,79.89) .. (461.16,81.46) .. controls (461.31,83.02) and (460.16,84.41) .. (458.6,84.56) .. controls (457.04,84.71) and (455.65,83.56) .. (455.5,82) -- cycle ;
    \draw  [color={rgb, 255:red, 65; green, 117; blue, 5 }  ,draw opacity=1 ][fill={rgb, 255:red, 0; green, 0; blue, 0 }  ,fill opacity=1 ] (440.5,92) .. controls (440.35,90.44) and (441.49,89.05) .. (443.06,88.9) .. controls (444.62,88.75) and (446.01,89.89) .. (446.16,91.46) .. controls (446.31,93.02) and (445.16,94.41) .. (443.6,94.56) .. controls (442.04,94.71) and (440.65,93.56) .. (440.5,92) -- cycle ;
    \draw  [color={rgb, 255:red, 65; green, 117; blue, 5 }  ,draw opacity=1 ][fill={rgb, 255:red, 0; green, 0; blue, 0 }  ,fill opacity=1 ] (449.84,102.54) .. controls (449.69,100.98) and (450.84,99.59) .. (452.4,99.44) .. controls (453.96,99.29) and (455.35,100.44) .. (455.5,102) .. controls (455.65,103.56) and (454.51,104.95) .. (452.94,105.1) .. controls (451.38,105.25) and (449.99,104.11) .. (449.84,102.54) -- cycle ;
    \draw  [color={rgb, 255:red, 65; green, 117; blue, 5 }  ,draw opacity=1 ][fill={rgb, 255:red, 0; green, 0; blue, 0 }  ,fill opacity=1 ] (464.5,111) .. controls (464.35,109.44) and (465.49,108.05) .. (467.06,107.9) .. controls (468.62,107.75) and (470.01,108.89) .. (470.16,110.46) .. controls (470.31,112.02) and (469.16,113.41) .. (467.6,113.56) .. controls (466.04,113.71) and (464.65,112.56) .. (464.5,111) -- cycle ;
    \draw  [color={rgb, 255:red, 65; green, 117; blue, 5 }  ,draw opacity=1 ][fill={rgb, 255:red, 0; green, 0; blue, 0 }  ,fill opacity=1 ] (439.94,135.1) .. controls (439.79,133.54) and (440.94,132.15) .. (442.5,132) .. controls (444.06,131.85) and (445.45,132.99) .. (445.6,134.56) .. controls (445.75,136.12) and (444.61,137.51) .. (443.04,137.66) .. controls (441.48,137.81) and (440.09,136.66) .. (439.94,135.1) -- cycle ;
    \draw  [color={rgb, 255:red, 65; green, 117; blue, 5 }  ,draw opacity=1 ][fill={rgb, 255:red, 0; green, 0; blue, 0 }  ,fill opacity=1 ] (455.5,127) .. controls (455.35,125.44) and (456.49,124.05) .. (458.06,123.9) .. controls (459.62,123.75) and (461.01,124.89) .. (461.16,126.46) .. controls (461.31,128.02) and (460.16,129.41) .. (458.6,129.56) .. controls (457.04,129.71) and (455.65,128.56) .. (455.5,127) -- cycle ;
    \draw  [color={rgb, 255:red, 65; green, 117; blue, 5 }  ,draw opacity=1 ][fill={rgb, 255:red, 0; green, 0; blue, 0 }  ,fill opacity=1 ] (445.5,148) .. controls (445.35,146.44) and (446.49,145.05) .. (448.06,144.9) .. controls (449.62,144.75) and (451.01,145.89) .. (451.16,147.46) .. controls (451.31,149.02) and (450.16,150.41) .. (448.6,150.56) .. controls (447.04,150.71) and (445.65,149.56) .. (445.5,148) -- cycle ;
    \draw  [color={rgb, 255:red, 65; green, 117; blue, 5 }  ,draw opacity=1 ][fill={rgb, 255:red, 0; green, 0; blue, 0 }  ,fill opacity=1 ] (462.5,156) .. controls (462.35,154.44) and (463.49,153.05) .. (465.06,152.9) .. controls (466.62,152.75) and (468.01,153.89) .. (468.16,155.46) .. controls (468.31,157.02) and (467.16,158.41) .. (465.6,158.56) .. controls (464.04,158.71) and (462.65,157.56) .. (462.5,156) -- cycle ;
    \draw  [color={rgb, 255:red, 65; green, 117; blue, 5 }  ,draw opacity=1 ][fill={rgb, 255:red, 0; green, 0; blue, 0 }  ,fill opacity=1 ] (465.5,172) .. controls (465.35,170.44) and (466.49,169.05) .. (468.06,168.9) .. controls (469.62,168.75) and (471.01,169.89) .. (471.16,171.46) .. controls (471.31,173.02) and (470.16,174.41) .. (468.6,174.56) .. controls (467.04,174.71) and (465.65,173.56) .. (465.5,172) -- cycle ;
    \draw [color={rgb, 255:red, 65; green, 117; blue, 5 }  ,draw opacity=1 ]   (575.5,60) .. controls (595.5,62) and (599.5,78) .. (577.5,83) .. controls (554.5,90) and (566.5,103) .. (577.5,103) .. controls (594.5,107) and (596.5,119) .. (577.5,128) .. controls (551.5,133) and (566.5,151) .. (579.5,153) .. controls (593.5,159) and (602.5,172) .. (577.5,179) ;
    \draw [shift={(577.5,179)}, rotate = 164.36] [color={rgb, 255:red, 65; green, 117; blue, 5 }  ,draw opacity=1 ][fill={rgb, 255:red, 65; green, 117; blue, 5 }  ,fill opacity=1 ][line width=0.75]      (0, 0) circle [x radius= 3.35, y radius= 3.35]   ;
    \draw [shift={(575.5,60)}, rotate = 5.71] [color={rgb, 255:red, 65; green, 117; blue, 5 }  ,draw opacity=1 ][fill={rgb, 255:red, 65; green, 117; blue, 5 }  ,fill opacity=1 ][line width=0.75]      (0, 0) circle [x radius= 3.35, y radius= 3.35]   ;
    \draw  [color={rgb, 255:red, 65; green, 117; blue, 5 }  ,draw opacity=1 ][fill={rgb, 255:red, 0; green, 0; blue, 0 }  ,fill opacity=1 ] (588.5,69) .. controls (588.35,67.44) and (589.49,66.05) .. (591.06,65.9) .. controls (592.62,65.75) and (594.01,66.89) .. (594.16,68.46) .. controls (594.31,70.02) and (593.16,71.41) .. (591.6,71.56) .. controls (590.04,71.71) and (588.65,70.56) .. (588.5,69) -- cycle ;
    \draw  [color={rgb, 255:red, 65; green, 117; blue, 5 }  ,draw opacity=1 ][fill={rgb, 255:red, 0; green, 0; blue, 0 }  ,fill opacity=1 ] (562.5,93) .. controls (562.35,91.44) and (563.49,90.05) .. (565.06,89.9) .. controls (566.62,89.75) and (568.01,90.89) .. (568.16,92.46) .. controls (568.31,94.02) and (567.16,95.41) .. (565.6,95.56) .. controls (564.04,95.71) and (562.65,94.56) .. (562.5,93) -- cycle ;
    \draw  [color={rgb, 255:red, 65; green, 117; blue, 5 }  ,draw opacity=1 ][fill={rgb, 255:red, 0; green, 0; blue, 0 }  ,fill opacity=1 ] (586.5,112) .. controls (586.35,110.44) and (587.49,109.05) .. (589.06,108.9) .. controls (590.62,108.75) and (592.01,109.89) .. (592.16,111.46) .. controls (592.31,113.02) and (591.16,114.41) .. (589.6,114.56) .. controls (588.04,114.71) and (586.65,113.56) .. (586.5,112) -- cycle ;
    \draw  [color={rgb, 255:red, 65; green, 117; blue, 5 }  ,draw opacity=1 ][fill={rgb, 255:red, 0; green, 0; blue, 0 }  ,fill opacity=1 ] (577.5,128) .. controls (577.35,126.44) and (578.49,125.05) .. (580.06,124.9) .. controls (581.62,124.75) and (583.01,125.89) .. (583.16,127.46) .. controls (583.31,129.02) and (582.16,130.41) .. (580.6,130.56) .. controls (579.04,130.71) and (577.65,129.56) .. (577.5,128) -- cycle ;
    \draw  [color={rgb, 255:red, 65; green, 117; blue, 5 }  ,draw opacity=1 ][fill={rgb, 255:red, 0; green, 0; blue, 0 }  ,fill opacity=1 ] (567.5,149) .. controls (567.35,147.44) and (568.49,146.05) .. (570.06,145.9) .. controls (571.62,145.75) and (573.01,146.89) .. (573.16,148.46) .. controls (573.31,150.02) and (572.16,151.41) .. (570.6,151.56) .. controls (569.04,151.71) and (567.65,150.56) .. (567.5,149) -- cycle ;
    \draw  [color={rgb, 255:red, 65; green, 117; blue, 5 }  ,draw opacity=1 ][fill={rgb, 255:red, 0; green, 0; blue, 0 }  ,fill opacity=1 ] (587.5,173) .. controls (587.35,171.44) and (588.49,170.05) .. (590.06,169.9) .. controls (591.62,169.75) and (593.01,170.89) .. (593.16,172.46) .. controls (593.31,174.02) and (592.16,175.41) .. (590.6,175.56) .. controls (589.04,175.71) and (587.65,174.56) .. (587.5,173) -- cycle ;
    \draw  [color={rgb, 255:red, 65; green, 117; blue, 5 }  ,draw opacity=1 ][fill={rgb, 255:red, 0; green, 0; blue, 0 }  ,fill opacity=1 ] (572.67,60.27) .. controls (572.52,58.71) and (573.67,57.32) .. (575.23,57.17) .. controls (576.79,57.02) and (578.18,58.17) .. (578.33,59.73) .. controls (578.48,61.29) and (577.33,62.68) .. (575.77,62.83) .. controls (574.21,62.98) and (572.82,61.83) .. (572.67,60.27) -- cycle ;
    \draw  [color={rgb, 255:red, 65; green, 117; blue, 5 }  ,draw opacity=1 ][fill={rgb, 255:red, 0; green, 0; blue, 0 }  ,fill opacity=1 ] (451.5,176) .. controls (451.35,174.44) and (452.49,173.05) .. (454.06,172.9) .. controls (455.62,172.75) and (457.01,173.89) .. (457.16,175.46) .. controls (457.31,177.02) and (456.16,178.41) .. (454.6,178.56) .. controls (453.04,178.71) and (451.65,177.56) .. (451.5,176) -- cycle ;
    \draw  [fill={rgb, 255:red, 65; green, 117; blue, 5 }  ,fill opacity=1 ] (505,118.5) -- (520.9,118.5) -- (520.9,115) -- (531.5,122) -- (520.9,129) -- (520.9,125.5) -- (505,125.5) -- cycle ;
    \draw [color={rgb, 255:red, 65; green, 117; blue, 5 }  ,draw opacity=1 ] [dash pattern={on 0.84pt off 2.51pt}]  (578.33,59.73) .. controls (588.17,58) and (598.5,63) .. (591.6,71.56) ;

    \draw [color={rgb, 255:red, 65; green, 117; blue, 5 }  ,draw opacity=1 ] [dash pattern={on 0.84pt off 2.51pt}]  (591.6,71.56) .. controls (604.5,84) and (583.5,82) .. (568.16,92.46) ;

    \draw [color={rgb, 255:red, 65; green, 117; blue, 5 }  ,draw opacity=1 ] [dash pattern={on 0.84pt off 2.51pt}]  (565.6,95.56) .. controls (578.5,99) and (598.5,103) .. (592.16,111.46) ;

    \draw [color={rgb, 255:red, 65; green, 117; blue, 5 }  ,draw opacity=1 ] [dash pattern={on 0.84pt off 2.51pt}]  (592.16,111.46) .. controls (605.5,133) and (596.06,129.9) .. (583.16,127.46) ;

    \draw [color={rgb, 255:red, 65; green, 117; blue, 5 }  ,draw opacity=1 ] [dash pattern={on 0.84pt off 2.51pt}]  (580.6,130.56) .. controls (568.5,130) and (571.5,138) .. (570.06,145.9) ;

    \draw [color={rgb, 255:red, 65; green, 117; blue, 5 }  ,draw opacity=1 ] [dash pattern={on 0.84pt off 2.51pt}]  (573.16,148.46) .. controls (595.5,151) and (609.5,151) .. (593.16,172.46) ;

    \draw [color={rgb, 255:red, 65; green, 117; blue, 5 }  ,draw opacity=1 ] [dash pattern={on 0.84pt off 2.51pt}]  (590.6,175.56) .. controls (572.5,196) and (583.5,199) .. (588.5,201) ;

    \draw [color={rgb, 255:red, 65; green, 117; blue, 5 }  ,draw opacity=1 ] [dash pattern={on 0.84pt off 2.51pt}]  (575.77,62.83) .. controls (575.5,72) and (584.5,70) .. (565.06,89.9) ;

    \draw [color={rgb, 255:red, 65; green, 117; blue, 5 }  ,draw opacity=1 ] [dash pattern={on 0.84pt off 2.51pt}]  (562.5,93) .. controls (562.23,102.17) and (570.5,113) .. (586.5,112) ;

    \draw [color={rgb, 255:red, 65; green, 117; blue, 5 }  ,draw opacity=1 ] [dash pattern={on 0.84pt off 2.51pt}]  (589.33,111.73) .. controls (579.5,135) and (524.5,136) .. (567.5,149) ;

    \draw [color={rgb, 255:red, 65; green, 117; blue, 5 }  ,draw opacity=1 ] [dash pattern={on 0.84pt off 2.51pt}]  (570.6,151.56) .. controls (592.5,173) and (582.5,158) .. (590.06,169.9) ;

    \draw [color={rgb, 255:red, 65; green, 117; blue, 5 }  ,draw opacity=1 ] [dash pattern={on 0.84pt off 2.51pt}]  (587.5,173) .. controls (560.5,170) and (569.5,198) .. (561.5,203) ;

    \draw  (66.5,57) -- (66.5,202)(129,176) -- (29,176) (71.5,195) -- (66.5,202) -- (61.5,195) (122,171) -- (129,176) -- (122,181)  ;
    \draw    (88,45) -- (136.5,45) ;
    \draw [shift={(138.5,45)}, rotate = 180] [color={rgb, 255:red, 0; green, 0; blue, 0 }  ][line width=0.75]    (10.93,-3.29) .. controls (6.95,-1.4) and (3.31,-0.3) .. (0,0) .. controls (3.31,0.3) and (6.95,1.4) .. (10.93,3.29)   ;
    
    \draw    (494,201) -- (512.95,134.92) ;
    \draw [shift={(513.5,133)}, rotate = 466] [color={rgb, 255:red, 0; green, 0; blue, 0 }  ][line width=0.75]    (10.93,-3.29) .. controls (6.95,-1.4) and (3.31,-0.3) .. (0,0) .. controls (3.31,0.3) and (6.95,1.4) .. (10.93,3.29)   ;

    \draw (69.16,35.45) node  [align=left] {1};
    \draw (164.53,37.45) node  [align=left] {2};
    \draw (249.41,35.45) node  [align=left] {3};
    \draw (365.84,38.17) node  [align=left] {4};
    \draw (457.82,36.78) node  [align=left] {5};
    \draw (489,211) node  [align=left] {$\displaystyle z_{t} :Space-SP$};
    \draw (49,196) node [rotate=-90]  {$X_{t}$};
    \draw (111,197) node [rotate=-90]  {$Y_{t}$};
    \draw (111,31) node   {$h_{t}$};
    \draw (515,44) node   {$T$};

    \end{tikzpicture}
    \caption{\label{fig:snp_examp}}
\end{subfigure}
\caption{(a) Graphical Model of Eq. \eqref{eq:snp}. Figure adapted from \citep{Singh2019}. (b) An example application for the RNP, where we want to predict the rest of the frame at each time step of a video given some context pixels. This is just an example and does not display the real performance of the RNP. (c) An example application of the graphical model displayed in \ref{fig:snp_graphical_model}. We are given a sequence of fixed-length sample functions from a stochastic process slowly changing over time. The $z_{t}$ models the relations in the sample function but does not model the relations between the different time steps.}
\end{figure}
The graphical model of the generative process is shown in Fig. \ref{fig:snp_graphical_model}. In comparison to the NP in Eq. \ref{eq:cond_lat_mod} we add recurrence to the prior on $z$.
By following the above intuitive and straightforward steps, we end up with a new NP model family able to deal with stochastic processes changing over time.\\
Imagine the data points represent frames in a video, a sequence of images, as shown in Fig \ref{fig:snp_video}. For each frame, we are given some context points. Our goal is to fill in the remaining pixels for each frame.
This task accentuates the fact that the variables inside of the data points do not necessarily have a temporal dependency between each other. For example, in a frame of a video at a time step $t$, no pixel in that frame is generated before any other pixel in the same frame. The stochastic process $z_{t}$ of Eq. \eqref{eq:snp} is, therefore, keeping track of the spatial relationships between the pixels of the same frame at time step $t$. The stochastic process $z_{t}$ is not modeling the temporal relationships of pixels between different time steps. It is the recurrent state that captures all the temporal relationships between pixels at different time steps.\\
In theory, the recurrent state of Eq. \eqref{eq:snp} can model all necessary dynamics. Also, in practice, the model shows good performance \citep{Singh2019}. However, the model from Eq. \eqref{eq:snp} lacks expressivity. It is not bringing the spirit of Neural Processes to sequences. It merely uses a recurrent state to take care of the sequence dynamics and uses the stochastic process for the already known problem domain. A model true to the idea of NPs would make use of the stochastic process also for modeling the temporal dynamics. We want to add a temporal stochastic process $v: \{z_{t}\}_{t\in T}$ that affects all of the spatial stochastic processes $z_{t}$, as motivated in Fig. \ref{fig:rnp_aligned}. This leads to a new generative process
\begin{equation}
    P ( \bold{Y} , V , Z | \bold{X} , C , T , L ) = \prod _ { t = 1 } ^ { T } \left( \prod _ { i \in I \left( D _ { t } \right) } P \left( \bold{y} _ {\bold{x}, i } ^ { t } | \bold{x} _ { i } ^ { t } , z _ { t } \right) \right) P \left( z _ { t } | z _ { < t } , C _ { t } , v , t \right) P ( v | L )\label{eq:rnp_rec}
\end{equation}
where $L = \left( T _ { L }, Z _ { L } \right) = \left( t , z _ { t } \right) _ { t \in I ( L ) }$ is the context for the temporal stochastic process $v$. We infer the temporal context points $z_{t}$ in $L$ from the spatial context points in $C_{t}$. The graphical model is shown in Fig. \ref{fig:rnp_graph_model1}. Since the temporal stochastic process $v$ is now capturing the changing dynamics, the recurrence over the spatial SP $z_{t}$ in Equation \eqref{eq:rnp_rec} might not be necessary. Leading to:
\begin{equation}
    P ( \bold{Y} , V , Z | \bold{X} , C , T , L ) = \prod _ { t = 1 } ^ { T } \left( \prod _ { i \in I \left( D _ { t } \right) } P \left( \bold{y} _ {\bold{x}, i } ^ { t } | \bold{x} _ { i ^ { \prime } } ^ { t } , z _ { t } \right) \right) P \left( z _ { t } | C _ { t } , v , t \right) P ( v | L )\label{eq:snp_norec}
\end{equation}
The corresponding graphical model is shown in Fig. \ref{fig:rnp_no_rec}.
\begin{figure}[]
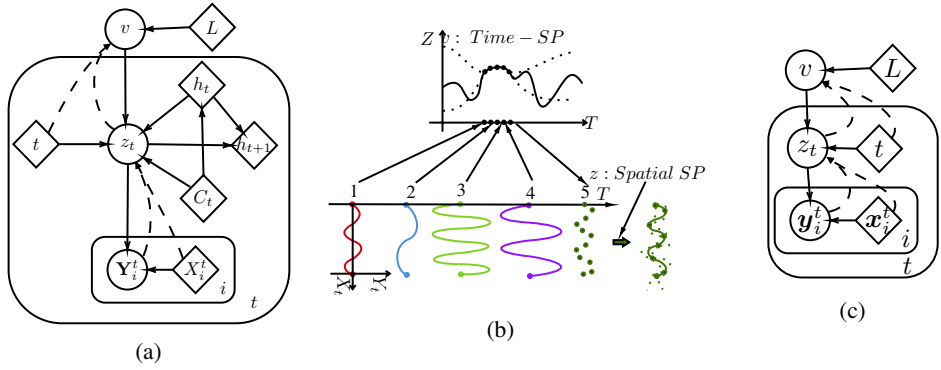

\begin{subfigure}[]{.33\textwidth}
\centering
    \tikzset{every picture/.style={line width=0.75pt}} 

    \caption{\label{fig:rnp_graph_model1}}
\end{subfigure}
\begin{subfigure}[]{.33\textwidth}
\tikzset{every picture/.style={line width=0.75pt}} 
%
 \caption{\label{fig:rnp_aligned}}
\end{subfigure}
\begin{subfigure}[]{.33\textwidth}
    \centering
    \tikzset{every picture/.style={line width=0.75pt}} 
    %
    \caption{ \label{fig:rnp_no_rec}}
\end{subfigure}
\caption{(a) Graphical model of the RNP with a temporal and spatial stochastic process. In this case, we have no recurrence over the temporal stochastic process. (b) For each data point, we infer its value in the temporal stochastic process. Here the observed timeline aligns perfectly with the latent timeline. This means that each $Z(t)$ parameterizes exactly one observed time step. We use the data points that we already know to infer the temporal SP. (c) In this graphical model, we have no recurrence over the spatial or the temporal SP. }
\end{figure}
\\ Eq.\eqref{eq:rnp_rec} and \eqref{eq:snp_norec} have a subtle aspect that turns out to be important for modelling sequences. The temporal stochastic process $v: \{z_{t}\}_{t \in T}$ \emph{shares} the same temporal indexing $T$ with our observed sequence of data points $\{(\boldsymbol{Y}^{t}_{i},\boldsymbol{X}_{i}^{t})\}_{i=1}^{N}\}_{t\in T}$, where $N$ is the size of the data points. That means latent and observed time steps are aligned; one input sequence $\{(\boldsymbol{Y}^{t}_{i},\boldsymbol{X}_{i}^{t})\}_{i=1}^{N}\}_{t}$ corresponds to one $z_{t}$. Clearly, this does not need to be the general case. For example, let us say the input sequences $\{(\boldsymbol{Y}^{t}_{i},\boldsymbol{X}_{i}^{t})\}_{i=1}^{N}\}_{t\in T}$ correspond to sine waves and $z_{t}$ corresponds to the amplitude of the sine wave. It is not unreasonable to assume that one $z_{t}$ might correspond to multiple input sequences. To differentiate between the observed time and the latent time, we introduce a new time line $T_{obs}$, which is the temporal indexing of the observed data $\{(\boldsymbol{Y}^{t_{obs}}_{i},\boldsymbol{X}_{i}^{t_{obs}})\}_{i=1}^{N}\}_{t\in T_{obs}}$. The time line $T$ stays the time line of the temporal stochastic process $v: \{z_{t}\}_{t \in T}$.
In general, we do not know how $T$ and $T_{obs}$ relate to each other. As a consequence, we have to find a mapping between the different indices $T \Longleftrightarrow T_{obs}$ to properly assign the credit of influence. We propose to do so via implicitly learning the connections in a bipartite graph $Q$ between two index sets. One set is formed by $T_{obs}$ and the other by $T$. The connections in-between the two disjoint sets have to satisfy the temporal causal constraints that when $t$ maps to $t_{obs}$, $t+1$ cannot map to $t^{\prime}_{obs}\leq t_{obs}$.
\begin{equation}
    P(Q\mid T,T_{obs}) = \prod_{t\in T}\prod_{t_{obs}\in T_{obs}}P(Q_{t,t_{obs}}\mid t_{obs},t)\label{eq:map}
\end{equation}
We can thus write the model as follows
\begin{align}
    P ( \bold{Y} , V , Z , T | \bold{X}, T_{obs} , C ) = &\prod _ { t_{obs} \in T_{obs} } \prod _ { t \in T }  \prod _ { i \in I \left( D _ { t_{obs} } \right) }\\ &P \left( \bold{y} _ { \bold{x}, i } ^ { t_{obs} } | \bold{x} _ { i ^ { \prime } } ^ { t_{obs} } , z _ { t } \right)P \left( z _ { t } | z _ { < t } , C _ { t_{obs} } , v , t \right) P ( Q | t_{obs} , t ) P ( v , t | L )\nonumber\label{eq:rnp_sing_lat}
\end{align}
One can imagine the model with or without recurrence at this point. The model is displayed in Figure \ref{fig:rnp_map}. Interestingly we can keep the recurrence on the spatial stochastic process. The updates happen at the frequency of the corresponding latent time $T$ and not $T_{obs}$. This is similar to other previously proposed methods where updates happen at different discrete time steps for different units in RNNs \citep{schmidhuber:chunkers,Schmidhuber:92ncchunker,el1996hierarchical,koutnik2014clockwork}.\\
Eq. \eqref{eq:rnp_sing_lat} only accounts for one latent time scale. Again, this might be overly restrictive: In the temperature example, we would want to model multiple times. The daily temperature fluctuations tick on a different time scale than the yearly seasons. In theory, it is possible for one temporal SP to account for these factors, but again, this lacks expressivity. This leads to the generative model
\begin{align}
    P ( \bold{Y} , V , Z, T| \bold{X} ,T_{obs}, C ) = &\prod _ { t_{obs} \in T_{obs} }\prod _ {\beta \in \mathcal{V} } \prod _ { i \in I ( D_{t_{obs}} ) } \prod _ { t \in T^{\beta}  }P \left( \bold{y} _ {\bold{x}, i } ^ { t_{obs} } | \bold{x} _ { i } ^ { t_{obs} } , z _ { t } ^ { \beta } \right)\\ &P \left( z _ { t } ^ { \beta } | z _ { < t } ^ { \beta } , C _ { t_{obs} } , v ^ { \beta } , t ^ { \beta } \right)P \left( Q ^ { \beta } | t ^ { \beta } , t_{obs} \right) P \left( v ^ { \beta } , t ^ { \beta } | L ^ { \beta } \right)\nonumber
\end{align}
with the graphical model shown in Figure \ref{fig:rnp_mult_lat} and a visualization shown in Fig. \ref{fig:intuit_rnp_mult_lat}.
\begin{figure}[]
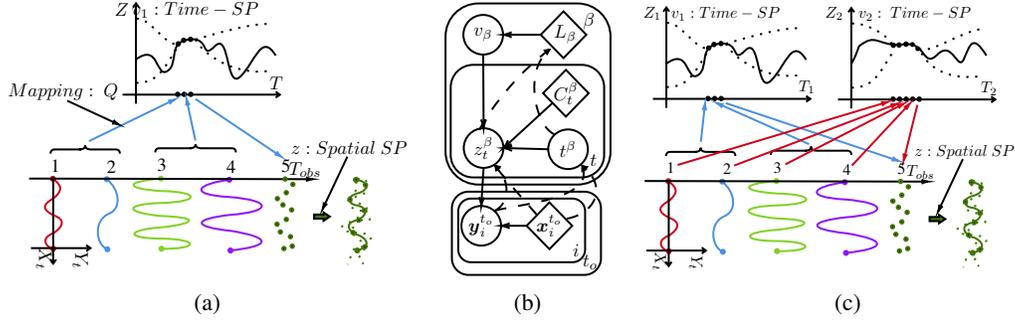

\begin{subfigure}[t]{.4\textwidth}
    \centering
    \tikzset{every picture/.style={line width=0.75pt}} 

    \caption{\label{fig:rnp_map}}
\end{subfigure}
\begin{subfigure}[t]{.2\textwidth}
\centering
    \tikzset{every picture/.style={line width=0.75pt}} 
    %
    \caption{\label{fig:rnp_mult_lat}}
\end{subfigure}
\begin{subfigure}[t]{.4\textwidth}
\tikzset{every picture/.style={line width=0.75pt}} 
%
    \caption{\label{fig:intuit_rnp_mult_lat}}
\end{subfigure}
\caption{(a) We map two time steps of the observed time to one time step of the latent time. The mapping is indicated by the blue arrows. (b) The graphical model of the RNP accounting for multiple temporal SPs. By introducing multiple temporal SPs, we explicitly model for different latent timelines. (c) Visualization of the RNP with multiple temporal SPs.}
\end{figure}
Finally, we arrived at the RNP. We were able to extend the NP to arrive at the RNP and thus show a generalization of the former. Now we will highlight some important aspects of the RNP. Then we proceed with presenting the architecture and a suitable loss function to train the model.\\
Whereas Eq. \eqref{eq:snp} only spans the spatial dimension with a stochastic process, the RNP additionally spans an SP over time. This has a few favorable consequences.
First, a temporal SP provides more options to induce biases about the problem, which can be explained with the following example. \\Consider Figure \ref{fig:hierarch_rnp}. You are given a series of temperature measurements, one measurement per hour for a few years. The data points are one-dimensional, so technically, the spatial stochastic process consists of only one variable, and the temporal stochastic process spans over the whole sequence. The temporal SP is, therefore, tasked to model all of the dynamics, and the spatial SP is idle.\\ However, given our prior knowledge about the temperature data, we can offload computation from the temporal SP to the spatial SP by splitting the time series into same-sized chunks, e.g., we split them into days. One chunk is considered one data point. Now the spatial SP is responsible for modeling the intra-day dynamics of temperature and the temporal SP of the inter-day dynamics.
\begin{figure}
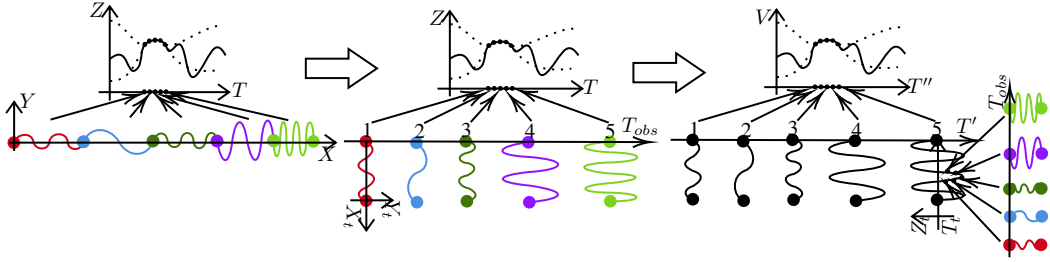

\centering
\tikzset{every picture/.style={line width=0.75pt}} 

\caption{ We can build a hierarchy of Neural Processes. In the temperature example, we could use one stochastic process to model the intra-day dynamics, another one to model the intra-week dynamics. The next stochastic process of the hierarchy could capture the intra-month dynamics, and so on.}
 \label{fig:hierarch_rnp}
\end{figure}
We cannot only chunk the observed sequence, but we can also chunk the temporal stochastic process and impose another stochastic process onto those chunks.  We could split the observed sequence into daily chunks, have a first level of temporal SP that models the intra-month dynamics, and a second level of temporal SP that models the inter-month dynamics and so on.  This leads to an architecture we term the Hierarchical Neural Process (HNP), which is the extension of Figure \ref{fig:rnp_no_rec} to an arbitrary number of SPs. If the chunks are self-similar, one could even think of a Recursive Neural Process. If we were to model recursive vegetables, we could even introduce the Romanesco Neural Process. Although the HNP was just derived from a time series example, it is not restricted to problems with a temporal structure.\\ The above application is also a good example to show another limitation of Eq. \eqref{eq:snp}. What if we had no intuition on how to split the temperature data? An example shown in Figure \ref{fig:application_rnp}. \\
The spatial SP of Equation \eqref{eq:snp} would be tasked with modeling only one value, whereas the recurrent hidden state would have to capture all the dynamics over time. In the RNP, all of the dynamics could be modeled by the temporal SP or recurrence. To have two components available to model the dynamics is advantageous. The temporal SP might be better suited to capture uncertainty than the recurrent state. The recurrent state can help to distinguish the true signal from noise.\\
The second consequence of a temporal SP is the ability to use information from future timesteps, if available, in a meaningful way. For example, imagine you are in a setup such as displayed in Figure \ref{fig:temp_sp_inbet2}. You are given information about your targeted time step from the past and from the future.\\
First, how would you include information from future timesteps in Eq. \eqref{eq:snp}? One could, for example, first process the available data points 1, 7, 18 and 26, appended with the time step, and then predict datapoint 13. The hidden state is then additionally tasked with keeping track of time jumps. This goes directly against the model design where we assume to receive data points in the sequence in which they were generated $\mathtt{P}_{t-1} \rightarrow \mathtt{P}_{t}$. Additionally, it is overloading the spatial stochastic process with uncertainty in time and space. However, in the RNP, the explicit indexing with time $T_{obs}$ enables a principled way of querying the temporal stochastic process for specific positions in time. This allows using future time steps as inducing points. Therefore, when we predict timestep 13 in Figure \ref{fig:temp_sp_inbet2}, information from the inducing points 18 and 26 of the temporal SP flows into the current prediction.\\
Now we proceed with the architectural details of the RNP.
\begin{figure}
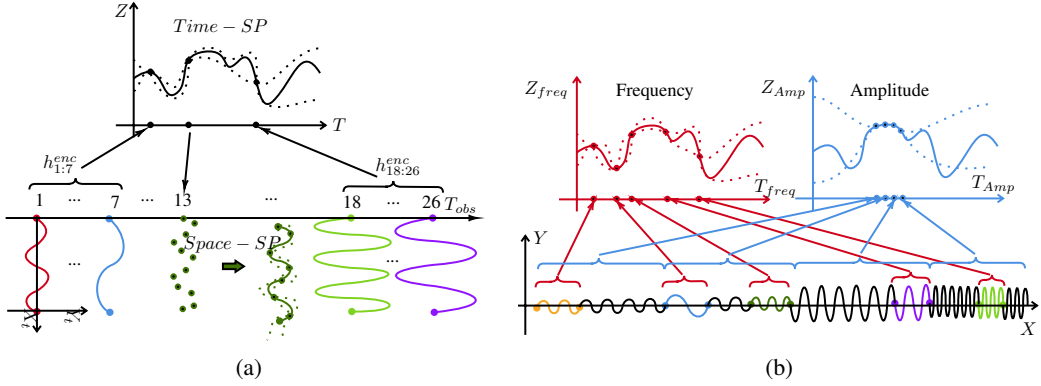

    \begin{subfigure}[t]{.5\textwidth}
    \centering
    \tikzset{every picture/.style={line width=0.75pt}} 

    \caption{\label{fig:temp_sp_inbet2}}
    \end{subfigure}
    \begin{subfigure}[t]{.5\textwidth}
    \centering
    \tikzset{every picture/.style={line width=0.75pt}} 
    %
    \caption{\label{fig:application_rnp}}
\end{subfigure}
    \caption{(a) If we are provided with observed time steps that are further apart, intuitively this should give better coverage of the temporal stochastic process. The idea of temporal SP also allows by design to account for future time steps. (b) If we do not know when the dynamics change or we have no prior knowledge about the problem setting, we have to deal with the one-dimensional time series. In this case, the spatial SP is underutilized. For example, in a sine wave we know it is parameterized through, e.g. Frequency and Amplitude. The RNP models the dynamics of the Frequency and Amplitude with a stochastic process inferred from previously seen time steps.}
    \label{fig:my_label}
\end{figure}
\subsection{Architecture}
We are given a sequence of one-dimensional data points, as displayed in Fig. \ref{fig:application_rnp} and we infer temporal stochastic processes by selecting a set of N subsequences $\{\{(x_{j},y_{j}),...,(x_{j+J},y_{j+J})\}^{i}\}_{i=1}^{N}$. The $J$ is the length of the subsequences. The index $j$ is the starting point of the subsequence and can be any legal value given the length of the time series. The selection of the subsequences is problem dependent. It is a convenient way of introducing prior knowledge about the problem into the model. The subsequences do not necessarily need to have the same length $J$. The formulation of the mapping allows variable length subsequences. For example, in Fig. \ref{fig:sfig1}, we select subsequences as contexts (on the left) and predict our target sequence (on the right).
\\ The subsequences are processed by an LSTM, which will result in a sequence of hidden states $\{\{\{h_{j}^{i}\}_{j=1}^{J}\}^{i}\} = f^{\text{LSTM\_enc}}(\{(x_{j},y_{j}),...,(x_{j+J},y_{j+J})\}^{i})$. The hidden states are used to produce a representation $z_{i} = f^{\text{MLP}}(h_{J}^{i})$ of the subsequence. There are many options on how to produce the representation. One could, for example, simply take the last hidden state $h_{J}^{i}$ as a representation of the sequence. The representations $z_{i}$ are aggregated, $\text{Agg}(z_{i}) = z$. A temporal SP $v$ is sampled, $v \sim \mathcal{N}(\mu_{v},\sigma_{v}), \text{ where } \left[\mu_{v},\sigma_{v}\right]=f^{v}(z)$. Since we do not have multidimensional inputs, there is no spatial SP, which means there are no spatial context points $C_{*}$. So we can proceed with predicting our target value $y^{*}$ at $x^{*}$ with our decoder: $y_{*}\mid v,x_{*},h_{t-1} \sim \mathcal{N}(\mu_{y},\sigma_{y}), \text{ where } \left[\mu_{y},\sigma_{y}\right]=f^{\text{LSTM\_dec}}(h_{t-1},v,x_{*})$. The decoder could be either an MLP or an LSTM, because the temporal stochastic process $v$ should be able to model all of the dynamics. We can even one-shot predict a whole sequence of length $m$ if we are certain about the governing dynamics of the next $m$ time steps. The architecture is visually described in the Appendix in Fig. \ref{fig:relevant_arch}.\\
This is the problem setting we are interested in for our experiments. That is why we provide a loss function for this model.
\subsection{ELBO}
We now proceed with deriving the ELBO for this specific RNP model. Note that we use indexing for both $z$ and $t$, although $t$ is already the indexing of $z$. This is done to stay consistent with previous literature, and it makes it clearer to distinguish between the context points and the target points. However, note that the index of $t$ represents the value that $t$ assumes.\\
Let $q(v\mid z_{1:L}, t_{1:L})$ be the variational posterior with which we can learn the non-linear decoder $g(\cdot)$. The Evidence Lower Bound then looks as follows:
\begin{equation}
    \text{log }p(z_{1:L}\mid t_{1:L}) \geq \mathbb{E}_{q\left(v | z_{1 : L}, t_{1 : L}\right)}\left[\sum_{i=1}^{L} \log p\left(z_{i} | v, t_{i}\right)+\log \frac{p(v)}{q\left(v | z_{1 : L}, t_{1 : L}\right)}\right]
\end{equation}
Equivalently to the NPs we split the dataset into a context set $z_{1:k}$,$t_{1:k}$ and a target set $z_{k+1:L}$,$t_{k+1:L}$
\begin{equation}
    \text{log }p(z_{k+1:L}\mid t_{1:L},z_{1:k}) \geq \mathbb{E}_{q\left(v | z_{1 : L}, t_{1 : L}\right)}\left[\sum_{i=k+1}^{L} \log p\left(z_{i} | v, t_{i}\right)+\log \frac{p (v  \mid  z_{1 : k}, t_{1 : k})}{q\left(v \mid z_{1 : L}, t_{1 : L}\right)}\right]
\end{equation}
As $p (v  \mid  z_{1 : k}, t_{1 : k})$ is intractable we will use $q (v  \mid  z_{1 : k}, t_{1 : k})$
\begin{equation}
    \text{log }p(z_{k+1:L}\mid t_{1:L},z_{1:k}) \geq \mathbb{E}_{q\left(v | z_{1 : L}, t_{1 : L}\right)}\left[\sum_{i=k+1}^{L} \log p\left(z_{i} | v, t_{i}\right)+\log \frac{q (v  \mid  z_{1 : k}, t_{1 : k})}{q\left(v \mid z_{1 : L}, t_{1 : L}\right)}\right]\label{eq:elbo_lat}
\end{equation}
Since we cannot directly observe the latent random functions we have to express this in terms of the observed data. We can express it with the observed data because we have a mapping $Q$ from the observed time steps and the latent time steps. Thanks to the mapping $Q$ we can approximate $(z_{1 : k}, t_{1 : k})$ with $(\bold{y}_{\{\bold{x} \in Q_{t_{1:k}}\}}, Q_{t_{1:k}})$, where $Q_{t_{1:k}} = \{\bold{x} \mid z \text{ at } t \text{ parameterizes } \bold{y} \text{ at } \bold{x}\}$. Basically the latent sequence provides a factorization for the joint distribution of the observed sequence. In other words, we have independent subsequences in the observed sequence data conditioned on the hidden sequence.
\begin{align}
    & \text{log }p(\bold{y}_{\{\bold{x} \in Q_{t_{k+1:L}}\}}\mid Q_{t_{1:L}}, \bold{y}_{\{\bold{x} \in Q_{t_{1:k}}\}})\label{eq:elbo_obs}\\ & \geq \mathbb{E}_{q\left(v | \bold{y}_{\{\bold{x} \in Q_{ t_{1:L}}\}}\right)}\left[\sum_{j=k+1}^{L}\left[\sum_{\bold{x}=\text{min}\{\bold{x} \in Q_{t_{j}}\}}^{\text{max}\{\bold{x} \in Q_{t_{j}}\}} \log p\left(\bold{y}_{\bold{x}} |\bold{x}, v\right)\right]+\log \frac{q (v  \mid  \bold{y}_{\{\bold{x} \in Q_{t_{1:k}}\}}, Q_{t_{1:k}})}{q\left(v \mid \bold{y}_{\{\bold{x} \in Q_{t_{1:L}}\}}, Q_{t_{1:L}}\right)}\right]\nonumber
\end{align}
The jump from Eq. \eqref{eq:elbo_lat} to Eq. \eqref{eq:elbo_obs} is again justified via the mapping.
Basically, we approximate $(z_{1 : k}, t_{1 : k})$ with $(\bold{y}_{\{\bold{x} \in Q_{t_{1:k}}\}}, Q_{t_{1:k}})$, where $Q_{t_{1:k}} = \{\bold{x} \mid z \text{ at } t \text{ parameterizes } \bold{y} \text{ at } \bold{x}\}$.\\
 The difficult aspect of Eq. \eqref{eq:elbo_obs} is that we also have to learn the mapping to approximate the right ELBO. As stated in Eq. \eqref{eq:elbo_obs}, we need to iterate over independent subsequences of our observed time series to calculate the correct loss. In theory, we would have to select the subsequences corresponding to the mapping to the hidden variables. In practice, however, we do not know what the mapping looks like. An option is to choose all possible subsequences. Another option is to select subsequences by hand.\\
\section{Experiments}
\textbf{Electricity}: The model is evaluated on the UCI Individual Household Electric Power Consumption Dataset \citep{Dua:2019}. This is a time series of 2'075'259 measurements collected at 1-minute intervals from a single household.  Each measurement consists of 
7 features related to power consumption. In order to compare to other works, we choose the feature ``active power'' as our target and the other features as inputs. We compare to earlier results \citep{Lim2019}.
\textbf{Drives}: The Coupled Electric Drives is a dataset for nonlinear system identification \citep{Wigren2010}. It consists of 500 time steps and was produced by two motors driving a pulley using a flexible belt, where the input is the sum of the voltages on the motors, and the target is the speed of the belt. We compare our model to the results reported earlier \citep{Mattos2015}.
\textbf{Metrics}: We evaluate model accuracy using the Mean Squared Error (\textbf{MSE}) between the predicted mean and the target value. The MSEs are normalized based on LSTM performance, which allows for comparing to previous work.
The prediction interval coverage probability (\textbf{PICP}) is a way of measuring the quality of the predicted uncertainty. We measure the performance on a 90\% prediction interval. It is defined as $\operatorname{PICP}=\frac{1}{T} \sum_{t=1}^{T} c_{t}$, where $c_{t} = 1$ if $\psi(0.05, t)<y_{t}<\psi(0.95, t)$ and $c_{t} = 0$ otherwise. $\psi(0.05, t)$ is the 5th percentile derived from predictions sampled from $N\left(f\left(\boldsymbol{x}_{t}\right), \mathbf{\sigma}\right)$ \citep{Lim2019}.\\
\section{Results}
\begin{figure}[]
\begin{subfigure}[t]{.5\textwidth}
  \centering
  \includegraphics[width=\linewidth]{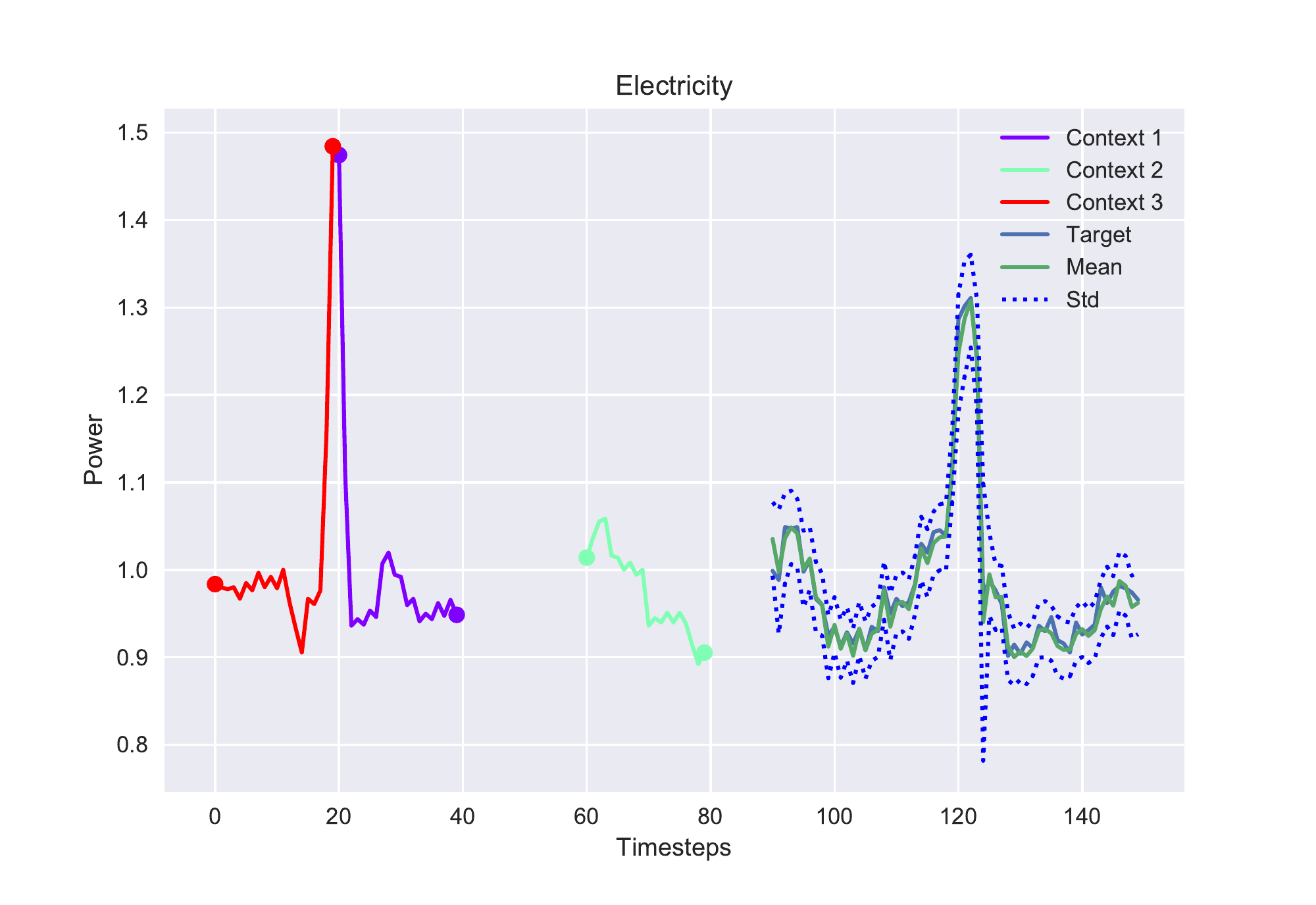}
  \caption{\label{fig:sfig1}}
\end{subfigure}%
\begin{subfigure}[t]{.5\textwidth}
  \centering
  \includegraphics[width=\linewidth]{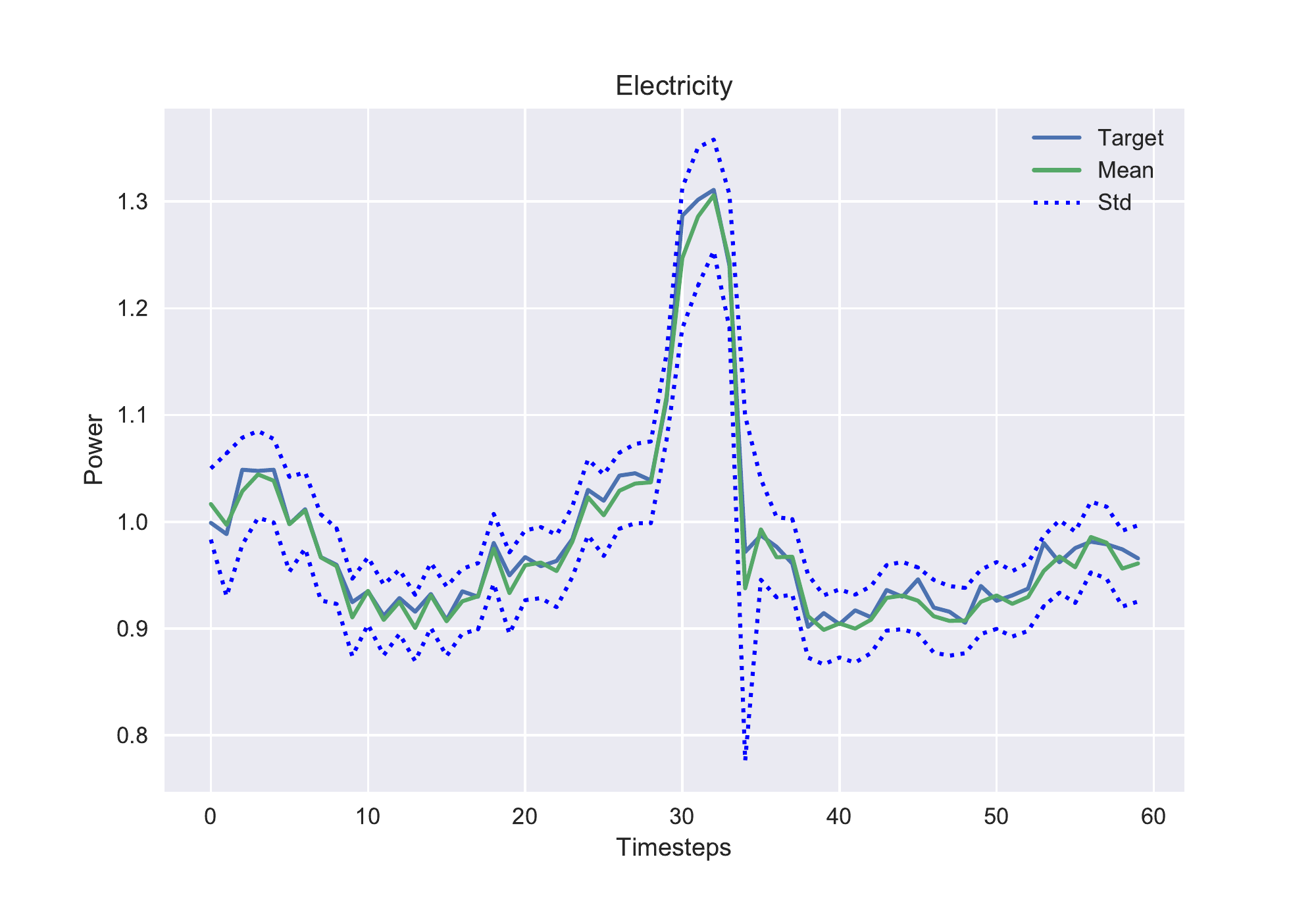}
  \caption{\label{fig:sfig4} }
\end{subfigure}
\begin{subfigure}[t]{.5\textwidth}
  \includegraphics[width=\linewidth]{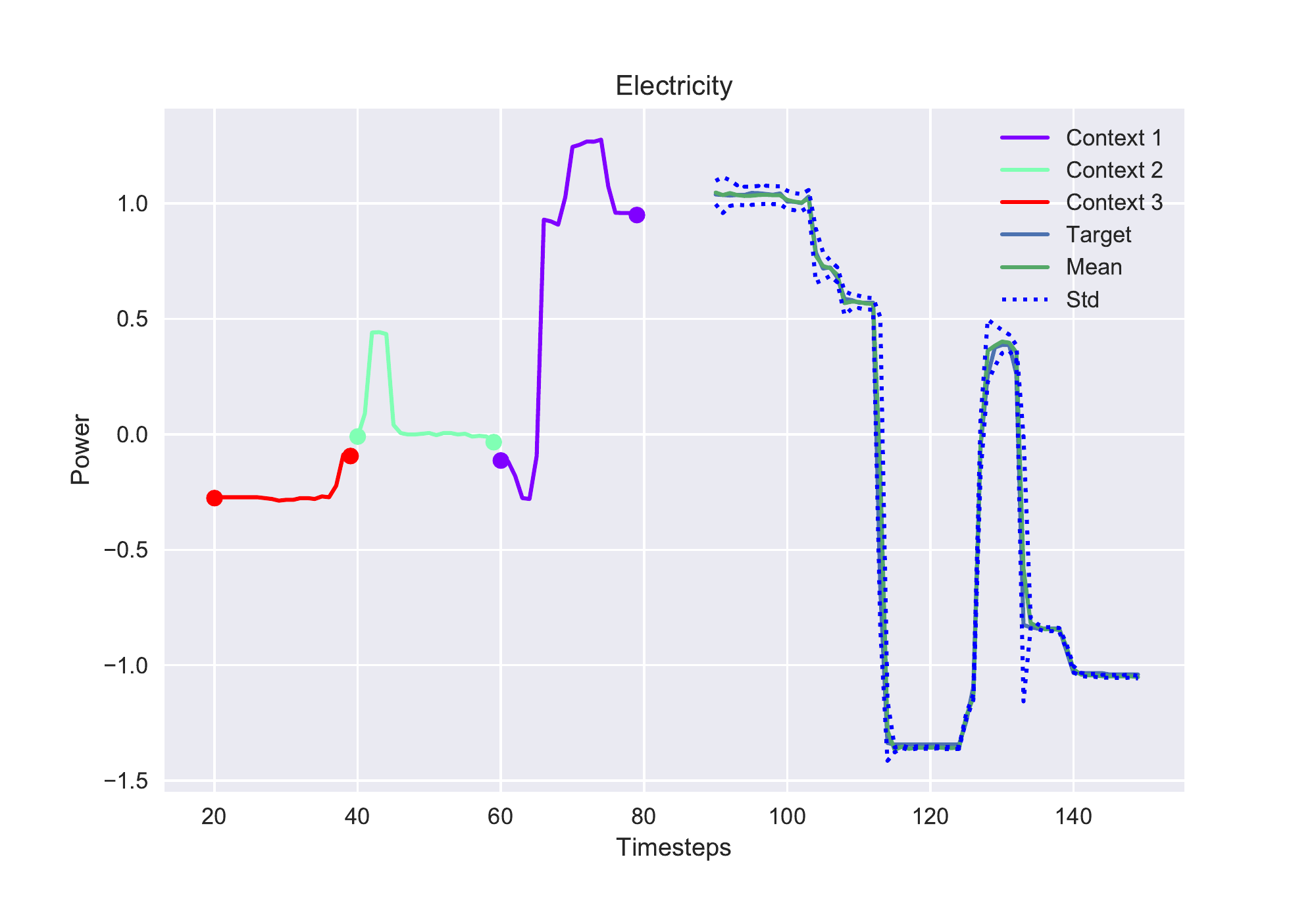}
  \caption{\label{fig:sfig2}}
\end{subfigure}
\begin{subfigure}[t]{.5\textwidth}
  \centering
  \includegraphics[width=\linewidth]{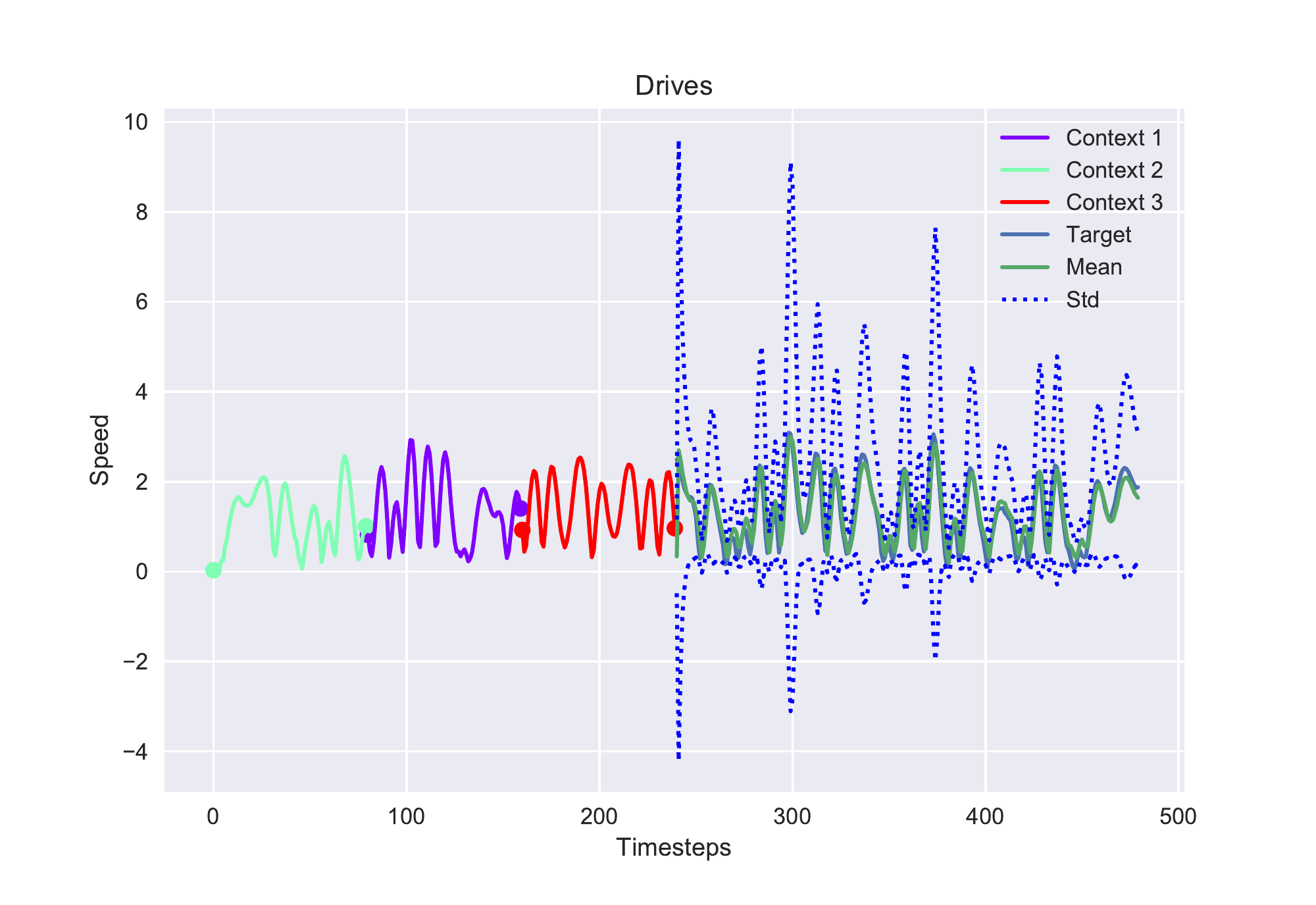}
  \caption{\label{fig:sfig3} }
\end{subfigure}
\caption{\label{fig:fig2} (a) A sample of the predictive performance on the Electricity dataset. The green line depicts the mean; the blue dotted line depicts one standard deviation. The blue line is the target. The multi-colored segments are randomly sampled subsequences used as context.(b) Close-up of the predicted sequence from the previous figure.(c) A second sample of the predictive performance on the Electricity dataset. The model is able to predict the start of the target sequence, given the context sequences. (d) The model is able to approximate the second half of the Drives dataset. However, it is relatively uncertain about the start point.}
\end{figure}

The results in Table \ref{table:mse} and Table \ref{table:picp} suggest that the performance of the RNP model is comparable to that of other SSMs. The outcome for the Electricity dataset indicates that the conditioning of the LSTM decoder was deceptive, rather than informative. This could be due to multiple factors (we did not choose an exhaustive amount of subsequences or sample them intelligently). For Drives, RNPs outperform LSTM and suggest informative uncertainty measures. For a qualitative analysis, see Figures \ref{fig:sfig1}-\ref{fig:sfig4}. The model learns to capture the time series. However, there is unwanted behavior. Figures \ref{fig:sfig1}-\ref{fig:sfig2} show that the RNP can predict the start of the target sequence and adapt its uncertainty.
\newpage
\section{Conclusion}
One significant benefit of Neural Processes is its improved computational cost over Gaussian Processes. This benefit becomes especially prominent when stacking the stochastic processes like in the Hierarchical Neural Process, which we hinted at in this thesis. The meta-learning background of the Neural Process also helps in time series analysis. Some regular events happen very rarely. The switch from spring to summer happens only once a year, or the crash of the stock market happens every decade or so. Since the Neural Process was designed to deal with a lack of data, these regular but rare events are, at least in theory, more easily captured. The good performance on the Drives dataset also indicates that the RNP framework can profit from the meta-learning background. \\
To conclude, we introduce a family of models called Recurrent Neural Processes (RNPs), a generalization of Neural Processes (NPs) to sequences by introducing a notion of latent time through modeling the state space with NPs, with a wide range of applications. RNPs can derive appropriate slow latent time scales from long sequences of quickly changing observations hiding specific long term patterns.\\
The framework derived for RNPs enables efficient inference of temporal context by establishing conditional independence among subsequences in a given time series.
It also provides an appropriate loss function for training RNPs. RNPs are also not restricted to modeling only two stochastic processes. The framework can be arbitrarily expanded to a hierarchy of stochastic processes, as shown in Fig. \ref{fig:hierarch_rnp}.
\begin{table}[]
\parbox{.45\linewidth}{
\centering
\caption{\label{table:mse} Normalized MSE for One-Step Predictions}
\begin{tabular}{llll}
\toprule
                & \textbf{Electricity} & \textbf{Drives} \\ \hline
\textbf{LSTM}   & 1.000                       & 1.000           \\ \hline
\textbf{VRNN}   & 1.902                          & -            \\
\textbf{DKF}    & 1.252                           & -            \\ \hline
\textbf{DSSM}   & 1.131                           & -            \\ \hline
\textbf{RNF-LG} & 0.918                           & -            \\
\textbf{RNF-NP} & 0.856                          & -            \\ \hline
\textbf{MLP-NARX} & -                         & 1.017            \\ \hline
\textbf{GP-NARX} & -                          & 0.953            \\ \hline
\textbf{REVARB} & -                        & 0.462            \\ \hline
\textbf{RNP}    & 1.111                         & 0.238          \\ \bottomrule
\end{tabular}

}
\hfill
\parbox{.45\linewidth}{
\centering
\caption{\label{table:picp} PICP for One-Step Prediction}
\begin{tabular}{llll}
\toprule
                & \textbf{Electricity}  & \textbf{Drives} \\ \hline
\textbf{VRNN}   & 0.986                          & -            \\
\textbf{DKF}    & 1.000                           & -            \\ \hline
\textbf{DSSM}   & 0.964                         & -            \\ \hline
\textbf{RNF-LG} & 0.960                          & -            \\
\textbf{RNF-NP} & 0.927                         & -            \\ \hline
\textbf{RNP}    & 0.947                          & 0.874            \\
\bottomrule
\end{tabular}

}
\end{table}

\subsubsection*{Acknowledgments}
We would like to thank Florian Trifterer, Pranav Shyam, Giorgio Giannone, Jan Eric Lenssen, David Ackermann and Heng Xin Fun for insightful discussions, and everyone at NNAISENSE for being part of such a conducive research environment.

\newpage
\bibliographystyle{plainnat}
\bibliography{biblio/master_thesis.bib}
\newpage
\section*{Appendix}
\subsection*{Overview}
The Supplementary Material presents more details about the RNP's implementation, the parameter search, and the experiments.
\subsection*{Experimental Details}
For all three experiments, we performed a coarse hyperparameter search in a  grid with the following possible value ranges. Minibatch size was always 256 except for the smaller Drives set, where it was 8. We trained the RNP using teacher forcing. The experiments were conducted on p3.16xlarge instances on the Amazon Web Services.
\begin{table}[H]
\centering
\parbox{.45\linewidth}{
\centering
\caption{\label{table:hyper} Parameter space searched during hyperparameter search.}
\begin{tabular}{llll}
\hline
                                    & \multicolumn{3}{l}{\textbf{Possible Values}} \\ \hline
\textbf{LSTM Hidden State Size}           & \multicolumn{3}{l}{3, 16, 32, (64)}                \\ 
\textbf{Context Sequence Length}    & \multicolumn{3}{l}{5, 50, 150 / 20, 60, 80}               \\
\textbf{Latent Vector Size}                & \multicolumn{3}{l}{4, 32, 128}               \\ 
\textbf{Bidirectional LSTM-Encoder} & \multicolumn{3}{l}{Yes, No}                  \\ 
\textbf{LSTM Layers}                & \multicolumn{3}{l}{1, 2}                     \\\hline
\end{tabular}}
\end{table}

To manage the experiments and ensure reproducibility we used Sacred \citep{greff2017}.

\subsubsection*{Electricity}
We reproduced the LSTM baseline model of previous work \citep{Lim2019} as accurately as possible, to compare our model to the reported normalized performances. 
For dataset preprocessing and architectural details of the other models we refer to the appendix of the aforementioned paper.\\ We tried multiple context sequence lengths, \{50, 150, 300\}, 
to train and test the network, but length did not have a significant impact on performance.
Experiments on this dataset profited from an increased hidden state and a decreased latent vector size. We trained the model for 130 epochs.

\subsubsection*{Drives}
For the Drives dataset, we relied on results reported by previous work \citep{Mattos2015}. Parameter search favored a small hidden state size and a larger latent vector size. Due to the dataset size, we trained the model on shorter context and target sequences of length 5 and 15 respectively. For testing, we used longer sequences to which the model was able to adapt. We trained it for 100'000 epochs.

\begin{figure}
    \begin{subfigure}[t]{.5\textwidth}
      \centering
      \includegraphics[width=\linewidth]{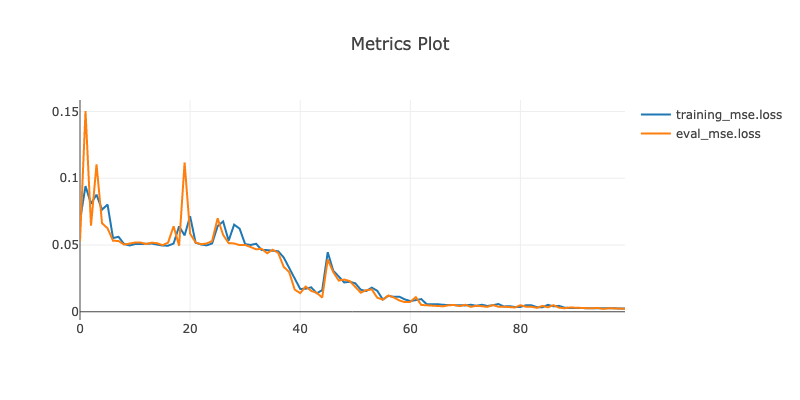}
      \caption{\label{fig:electricity} Learning curve of RNP on training and validation data of the Electricity dataset. Keep in mind that we do not optimize for MSE, but for log likelihood.}
    \end{subfigure}%
    \hspace{2mm}
    \begin{subfigure}[t]{.5\textwidth}
      \centering
      \includegraphics[width=\linewidth]{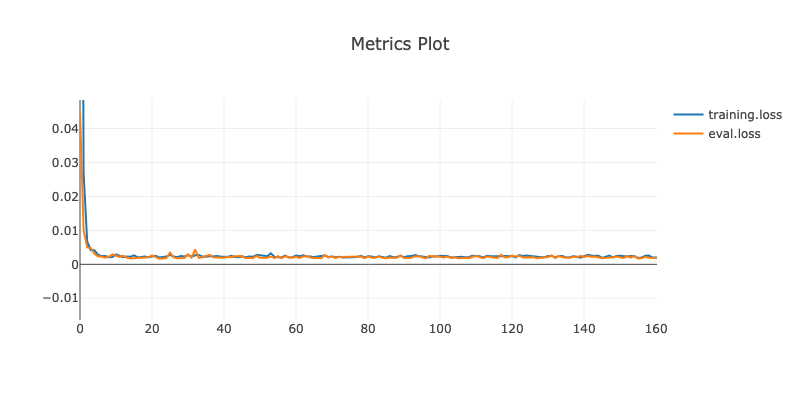}
      \caption{\label{fig:lstm_electricity} Learning curve of LSTM on training and validation data of the Electricity dataset, optimized for MSE.}
    \end{subfigure}%
    \\
    \begin{subfigure}[t]{.5\textwidth}
      \centering
      \includegraphics[width=\linewidth]{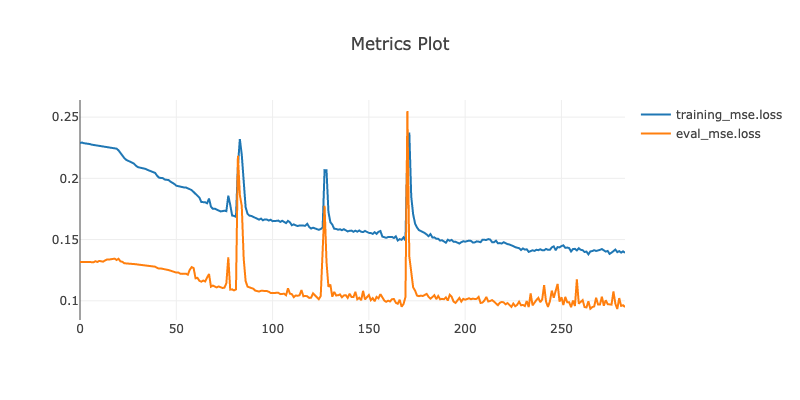}
      \caption{\label{fig:drives} Learning curve of RNP on training and validation data of the Drives dataset.}
    \end{subfigure}%
    \caption{Training and Validation Metrics for RNP and LSTM on the Electricity (\textbf{(a), (b)}) and Drives dataset (\textbf{(c)})}
\end{figure}

\begin{table}[H]
\caption{\label{table:hyper2} We used these hyperparameters for our final performance comparison.}
\begin{tabular}{lll|ll}
                                    & \multicolumn{2}{l}{\textbf{LSTM}} & \multicolumn{2}{l}{\textbf{RNP}} \\ \hline
                                       & Electricity   & Drives       & Electricity   & Drives  \\ \hline
\textbf{Hidden State Size}                    & 50            & up to 2048              & 64            & 32       \\
\textbf{Context Sequence Length}          & -             & -             & 20            & 80      \\
\textbf{Test Sequence Length}           & 50            & 250           & 60            & 250     \\
\textbf{Latent Vector Size}                      & -             & -              & 4             & 32     \\
\textbf{Bidirectional LSTM-Encoder}       & -             & -             & Yes           & No     \\
\textbf{LSTM Layers}                      & 1             & up to 3               & 2             & 2       \\
\textbf{Learning Rate}                  & 0.1           & -       & 0.01          & 0.0001  \\ \hline
\end{tabular}
\end{table}

\begin{figure}
    \begin{subfigure}[t]{.5\textwidth}
      \centering
      \includegraphics[width=\linewidth]{figures/elec_result_pdf.pdf}
      \caption{\label{fig:electricity} We can see that the network is able to capture the time series in its bounds of one standard deviation.}
    \end{subfigure}%
    \hspace{10mm}
    \begin{subfigure}[t]{.5\textwidth}
      \centering
      \includegraphics[width=\linewidth]{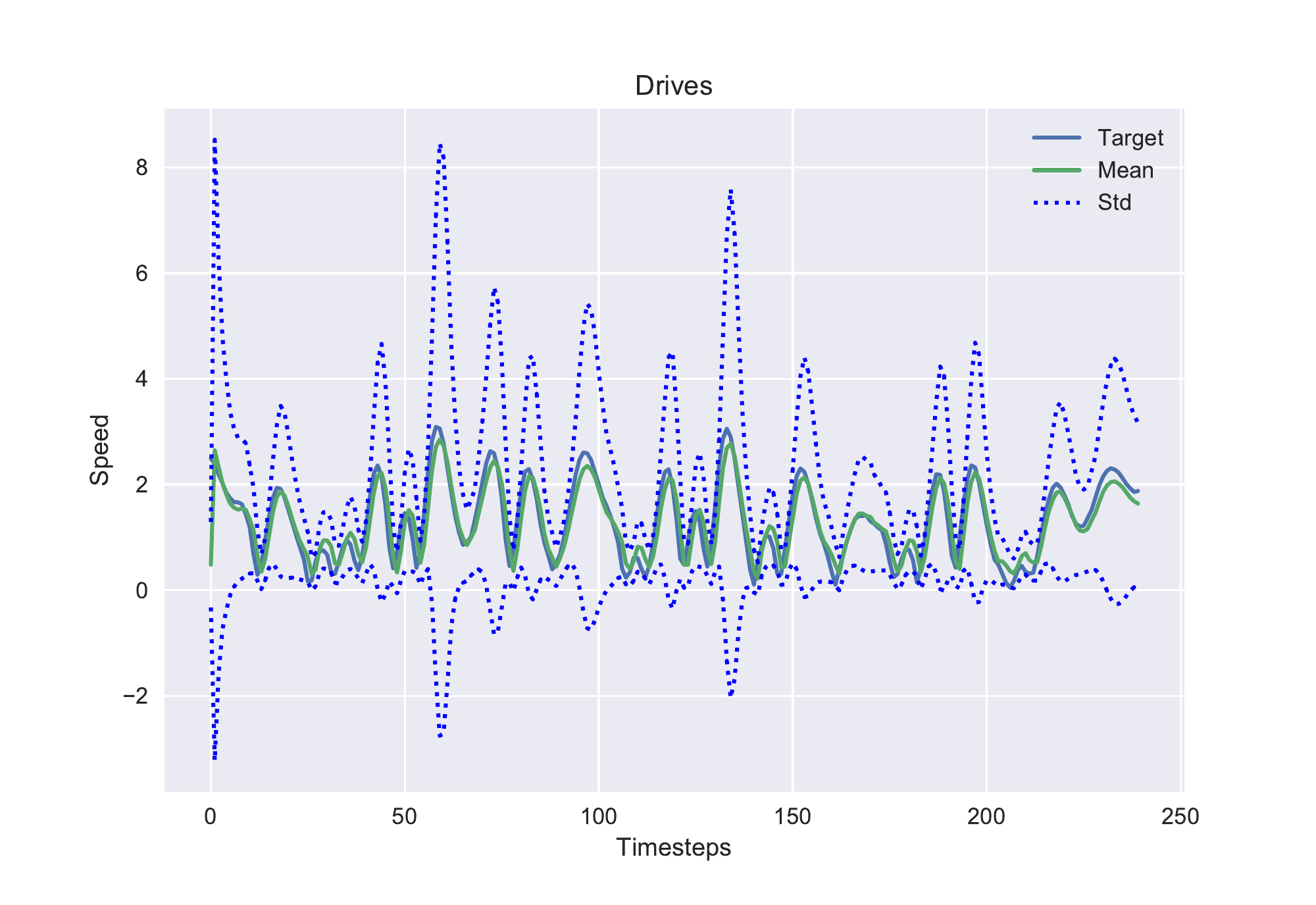}
      \caption{\label{fig:drives} The model captures the system dynamics; the apparent lag in prediction is possibly due to training by teacher forcing.}
    \end{subfigure}%
    \caption{Zooming in on RNP's predictive performance on Electricity (\textbf{(a)}) and Drives (\textbf{(b)})}
\end{figure}

\subsection*{Architecture}
\begin{figure}[H]
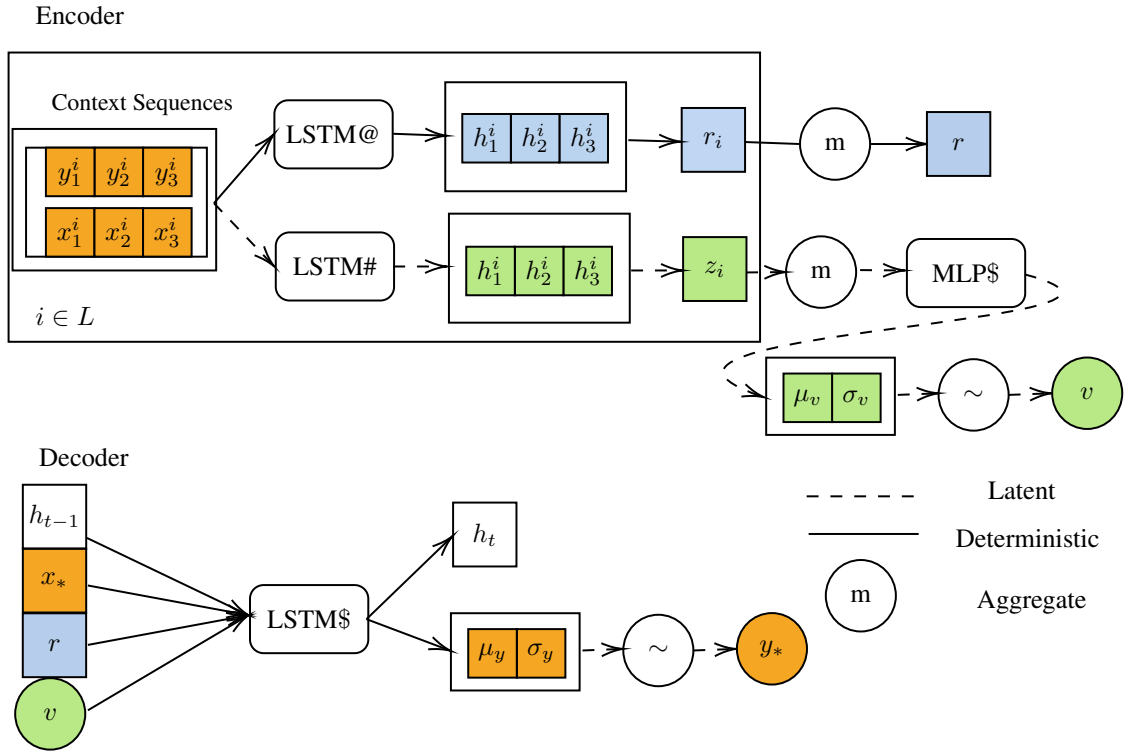

    \centering
    \tikzset{every picture/.style={line width=0.75pt}} 

    \caption{The architecture displayed here was used in the experiments, as explained in the Architecture section. We infer the temporal stochastic process from hand-selected subsequences. Adapted from \cite{Kim2019}}
    \label{fig:relevant_arch}
\end{figure}
The architecture depicted in Figure \ref{fig:relevant_arch} exhibits a stochastic and a deterministic path. Having both paths was reported to be beneficial \citep{Le2018}. Each path has its own encoder. Subsequences fed into the left will be encoded by LSTMs providing sequences of hidden states. 

Bidirectional paths yield two sequences of hidden states that one could encode in various ways. Our final model uses only the last hidden state and the corresponding input.  

The codes are aggregated and fed into the LSTM decoder at each time step. Inferring the first hidden state of the decoder from the representation was found to be helpful.
\end{document}